\pdfoutput=1
\documentclass[sigconf,nonacm]{acmart}

\usepackage{booktabs}
\usepackage{listings}
\usepackage{pifont}
\usepackage{tikz}
\usepackage{orcidlink}
\usepackage{fontawesome5}
\usetikzlibrary{arrows.meta,positioning,fit,backgrounds}

\newcommand{\code}[1]{\texttt{#1}}
\newcommand{\lib}{\textsf{torchtsetlin}}
\newcommand{\cmark}{\textcolor{green!55!black}{\ding{51}}}
\newcommand{\xmark}{\textcolor{red!75!black}{\ding{55}}}
\newcommand{\pmark}{$\circ$}
\providecommand{\R}[1]{\ifcsname R@#1\endcsname\csname R@#1\endcsname\else\textcolor{red}{??#1}\fi}
\expandafter\def\csname R@acc/B1/gt/default\endcsname{54.94}
\expandafter\def\csname R@acc/B1/pytm/default\endcsname{100.00}
\expandafter\def\csname R@acc/B1/pytmcuda/default\endcsname{100.00}
\expandafter\def\csname R@acc/B1/pytmpar/default\endcsname{100.00}
\expandafter\def\csname R@acc/B1/tmu/default\endcsname{100.00}
\expandafter\def\csname R@acc/B1/tt/b100\endcsname{100.00}
\expandafter\def\csname R@acc/B1/tt/batch\endcsname{100.00}
\expandafter\def\csname R@acc/B1/tt/seq\endcsname{100.00}
\expandafter\def\csname R@acc/B10/pytmcuda/default\endcsname{0.622}
\expandafter\def\csname R@acc/B10/tmu/default\endcsname{0.614}
\expandafter\def\csname R@acc/B10/tt/b100\endcsname{0.621}
\expandafter\def\csname R@acc/B10/tt/batch\endcsname{0.616}
\expandafter\def\csname R@acc/B10/tt/seq\endcsname{0.612}
\expandafter\def\csname R@acc/B2/pytm/default\endcsname{98.62}
\expandafter\def\csname R@acc/B2/pytmcuda/default\endcsname{98.82}
\expandafter\def\csname R@acc/B2/pytmpar/default\endcsname{99.97}
\expandafter\def\csname R@acc/B2/tmu/default\endcsname{99.73}
\expandafter\def\csname R@acc/B2/tt/b100\endcsname{92.47}
\expandafter\def\csname R@acc/B2/tt/batch\endcsname{99.61}
\expandafter\def\csname R@acc/B2/tt/seq\endcsname{99.60}
\expandafter\def\csname R@acc/B3/gt/default\endcsname{97.37}
\expandafter\def\csname R@acc/B3/pytm/default\endcsname{99.12}
\expandafter\def\csname R@acc/B3/pytmcuda/default\endcsname{97.50}
\expandafter\def\csname R@acc/B3/pytmpar/default\endcsname{98.25}
\expandafter\def\csname R@acc/B3/tmu/default\endcsname{97.32}
\expandafter\def\csname R@acc/B3/tt/b100\endcsname{96.86}
\expandafter\def\csname R@acc/B3/tt/batch\endcsname{97.42}
\expandafter\def\csname R@acc/B3/tt/seq\endcsname{97.49}
\expandafter\def\csname R@acc/B4/gt/default\endcsname{98.05}
\expandafter\def\csname R@acc/B4/pytm/default\endcsname{98.02}
\expandafter\def\csname R@acc/B4/pytmcuda/default\endcsname{98.21}
\expandafter\def\csname R@acc/B4/pytmpar/default\endcsname{98.08}
\expandafter\def\csname R@acc/B4/tmu/default\endcsname{97.97}
\expandafter\def\csname R@acc/B4/tt/b100\endcsname{97.54}
\expandafter\def\csname R@acc/B4/tt/batch\endcsname{97.90}
\expandafter\def\csname R@acc/B5/gt/default\endcsname{10.49}
\expandafter\def\csname R@acc/B5/pytm/default\endcsname{98.12}
\expandafter\def\csname R@acc/B5/pytmcuda/default\endcsname{98.11}
\expandafter\def\csname R@acc/B5/pytmpar/default\endcsname{98.07}
\expandafter\def\csname R@acc/B5/tmu/default\endcsname{98.17}
\expandafter\def\csname R@acc/B5/tt/b100\endcsname{98.14}
\expandafter\def\csname R@acc/B5/tt/batch\endcsname{98.15}
\expandafter\def\csname R@acc/B7/pytm/default\endcsname{99.12}
\expandafter\def\csname R@acc/B7/pytmcuda/default\endcsname{98.94}
\expandafter\def\csname R@acc/B7/pytmpar/default\endcsname{99.12}
\expandafter\def\csname R@acc/B7/tmu/default\endcsname{99.11}
\expandafter\def\csname R@acc/B7/tt/b100\endcsname{99.03}
\expandafter\def\csname R@acc/B7/tt/batch\endcsname{99.11}
\expandafter\def\csname R@acc/B8/pytmcuda/default\endcsname{88.37}
\expandafter\def\csname R@acc/B8/tmu/default\endcsname{89.97}
\expandafter\def\csname R@acc/B8/tt/b100\endcsname{89.70}
\expandafter\def\csname R@acc/B8/tt/batch\endcsname{90.06}
\expandafter\def\csname R@acc/B9/pytmcuda/default\endcsname{81.27}
\expandafter\def\csname R@acc/B9/tmu/default\endcsname{88.39}
\expandafter\def\csname R@acc/B9/tt/b100\endcsname{87.99}
\expandafter\def\csname R@acc/B9/tt/batch\endcsname{88.29}
\expandafter\def\csname R@bsix/gt-cpu1/10\endcsname{1.04}
\expandafter\def\csname R@bsix/gt-cpu1/100\endcsname{3.60}
\expandafter\def\csname R@bsix/gt-cpu1/1000\endcsname{35.0}
\expandafter\def\csname R@bsix/gt-cpu1/2000\endcsname{70.6}
\expandafter\def\csname R@bsix/gt-cpu1/5000\endcsname{177}
\expandafter\def\csname R@bsix/gt-cpu16/100\endcsname{29.4}
\expandafter\def\csname R@bsix/gt-cpu16/1000\endcsname{36.4}
\expandafter\def\csname R@bsix/gt-cpu16/2000\endcsname{45.7}
\expandafter\def\csname R@bsix/gt-cpu16/5000\endcsname{79.2}
\expandafter\def\csname R@bsix/gt-cpu4/10\endcsname{12.8}
\expandafter\def\csname R@bsix/gt-cpu4/100\endcsname{11.9}
\expandafter\def\csname R@bsix/gt-cpu4/1000\endcsname{26.0}
\expandafter\def\csname R@bsix/gt-cpu4/2000\endcsname{42.9}
\expandafter\def\csname R@bsix/gt-cpu4/5000\endcsname{97.1}
\expandafter\def\csname R@bsix/pytm-cpu1/10\endcsname{9.07}
\expandafter\def\csname R@bsix/pytm-cpu1/100\endcsname{20.1}
\expandafter\def\csname R@bsix/pytm-cpu1/1000\endcsname{95.8}
\expandafter\def\csname R@bsix/pytm-cpu1/2000\endcsname{174}
\expandafter\def\csname R@bsix/pytm-cpu1/5000\endcsname{422}
\expandafter\def\csname R@bsix/pytmcuda-cuda/10\endcsname{16.3}
\expandafter\def\csname R@bsix/pytmcuda-cuda/100\endcsname{27.7}
\expandafter\def\csname R@bsix/pytmcuda-cuda/1000\endcsname{31.3}
\expandafter\def\csname R@bsix/pytmcuda-cuda/2000\endcsname{37.7}
\expandafter\def\csname R@bsix/pytmcuda-cuda/5000\endcsname{90.2}
\expandafter\def\csname R@bsix/pytmpar-cpu16/10\endcsname{1.61}
\expandafter\def\csname R@bsix/pytmpar-cpu16/100\endcsname{2.64}
\expandafter\def\csname R@bsix/pytmpar-cpu16/1000\endcsname{11.8}
\expandafter\def\csname R@bsix/pytmpar-cpu16/2000\endcsname{36.9}
\expandafter\def\csname R@bsix/pytmpar-cpu16/5000\endcsname{115}
\expandafter\def\csname R@bsix/pytmpar-cpu4/10\endcsname{2.25}
\expandafter\def\csname R@bsix/pytmpar-cpu4/100\endcsname{5.48}
\expandafter\def\csname R@bsix/pytmpar-cpu4/1000\endcsname{37.1}
\expandafter\def\csname R@bsix/pytmpar-cpu4/2000\endcsname{97.3}
\expandafter\def\csname R@bsix/pytmpar-cpu4/5000\endcsname{262}
\expandafter\def\csname R@bsix/tmu-cpu1/10\endcsname{84.6}
\expandafter\def\csname R@bsix/tmu-cpu1/100\endcsname{101}
\expandafter\def\csname R@bsix/tmu-cpu1/1000\endcsname{262}
\expandafter\def\csname R@bsix/tmu-cpu1/2000\endcsname{432}
\expandafter\def\csname R@bsix/tmu-cpu1/5000\endcsname{916}
\expandafter\def\csname R@bsix/tmu-cuda/10\endcsname{177}
\expandafter\def\csname R@bsix/tmu-cuda/100\endcsname{146}
\expandafter\def\csname R@bsix/tmu-cuda/1000\endcsname{150}
\expandafter\def\csname R@bsix/tmu-cuda/2000\endcsname{158}
\expandafter\def\csname R@bsix/tmu-cuda/5000\endcsname{186}
\expandafter\def\csname R@bsix/tt-cuda[b100]/10\endcsname{1.93}
\expandafter\def\csname R@bsix/tt-cuda[b100]/100\endcsname{1.93}
\expandafter\def\csname R@bsix/tt-cuda[b100]/1000\endcsname{2.02}
\expandafter\def\csname R@bsix/tt-cuda[b100]/2000\endcsname{2.28}
\expandafter\def\csname R@bsix/tt-cuda[b100]/5000\endcsname{2.99}
\expandafter\def\csname R@bsix/tt-cuda[batch]/10\endcsname{3.61}
\expandafter\def\csname R@bsix/tt-cuda[batch]/100\endcsname{3.60}
\expandafter\def\csname R@bsix/tt-cuda[batch]/1000\endcsname{3.63}
\expandafter\def\csname R@bsix/tt-cuda[batch]/2000\endcsname{3.74}
\expandafter\def\csname R@bsix/tt-cuda[batch]/5000\endcsname{5.39}
\expandafter\def\csname R@count/seeds\endcsname{735}
\expandafter\def\csname R@count/timing\endcsname{228}
\expandafter\def\csname R@gpumib/B1/pytmcuda-cuda/default\endcsname{2{,}104}
\expandafter\def\csname R@gpumib/B1/tmu-cuda/default\endcsname{2{,}104}
\expandafter\def\csname R@gpumib/B1/tt-cuda/b100\endcsname{382}
\expandafter\def\csname R@gpumib/B1/tt-cuda/batch\endcsname{384}
\expandafter\def\csname R@gpumib/B1/tt-cuda/seq\endcsname{400}
\expandafter\def\csname R@gpumib/B10/pytmcuda-cuda/default\endcsname{2{,}104}
\expandafter\def\csname R@gpumib/B10/tmu-cuda/default\endcsname{2{,}104}
\expandafter\def\csname R@gpumib/B10/tt-cuda/b100\endcsname{438}
\expandafter\def\csname R@gpumib/B10/tt-cuda/batch\endcsname{438}
\expandafter\def\csname R@gpumib/B10/tt-cuda/seq\endcsname{458}
\expandafter\def\csname R@gpumib/B2/pytmcuda-cuda/default\endcsname{2{,}104}
\expandafter\def\csname R@gpumib/B2/tmu-cuda/default\endcsname{2{,}104}
\expandafter\def\csname R@gpumib/B2/tt-cuda/b100\endcsname{348}
\expandafter\def\csname R@gpumib/B2/tt-cuda/batch\endcsname{350}
\expandafter\def\csname R@gpumib/B2/tt-cuda/seq\endcsname{428}
\expandafter\def\csname R@gpumib/B3/pytmcuda-cuda/default\endcsname{2{,}104}
\expandafter\def\csname R@gpumib/B3/tmu-cuda/default\endcsname{2{,}104}
\expandafter\def\csname R@gpumib/B3/tt-cuda/b100\endcsname{388}
\expandafter\def\csname R@gpumib/B3/tt-cuda/batch\endcsname{386}
\expandafter\def\csname R@gpumib/B3/tt-cuda/seq\endcsname{396}
\expandafter\def\csname R@gpumib/B4/pytmcuda-cuda/default\endcsname{2{,}104}
\expandafter\def\csname R@gpumib/B4/tmu-cuda/default\endcsname{2{,}104}
\expandafter\def\csname R@gpumib/B4/tt-cuda/b100\endcsname{1{,}034}
\expandafter\def\csname R@gpumib/B4/tt-cuda/batch\endcsname{998}
\expandafter\def\csname R@gpumib/B5/pytmcuda-cuda/default\endcsname{2{,}104}
\expandafter\def\csname R@gpumib/B5/tmu-cuda/default\endcsname{2{,}104}
\expandafter\def\csname R@gpumib/B5/tt-cuda/b100\endcsname{1{,}036}
\expandafter\def\csname R@gpumib/B5/tt-cuda/batch\endcsname{1{,}000}
\expandafter\def\csname R@gpumib/B7/pytmcuda-cuda/default\endcsname{2{,}344}
\expandafter\def\csname R@gpumib/B7/tmu-cuda/default\endcsname{2{,}104}
\expandafter\def\csname R@gpumib/B7/tt-cuda/b100\endcsname{980}
\expandafter\def\csname R@gpumib/B7/tt-cuda/batch\endcsname{824}
\expandafter\def\csname R@gpumib/B8/pytmcuda-cuda/default\endcsname{2{,}344}
\expandafter\def\csname R@gpumib/B8/tmu-cuda/default\endcsname{2{,}104}
\expandafter\def\csname R@gpumib/B8/tt-cuda/b100\endcsname{980}
\expandafter\def\csname R@gpumib/B8/tt-cuda/batch\endcsname{824}
\expandafter\def\csname R@gpumib/B9/pytmcuda-cuda/default\endcsname{2{,}104}
\expandafter\def\csname R@gpumib/B9/tmu-cuda/default\endcsname{2{,}104}
\expandafter\def\csname R@gpumib/B9/tt-cuda/b100\endcsname{3{,}396}
\expandafter\def\csname R@gpumib/B9/tt-cuda/batch\endcsname{3{,}378}
\expandafter\def\csname R@repro/R1/acc\endcsname{98.16}
\expandafter\def\csname R@repro/R1/xtrain\endcsname{0.95}
\expandafter\def\csname R@repro/R2/acc\endcsname{98.16}
\expandafter\def\csname R@repro/R2/xtrain\endcsname{2.60}
\expandafter\def\csname R@repro/R3/acc\endcsname{90.70}
\expandafter\def\csname R@repro/R3/xtrain\endcsname{0.96}
\expandafter\def\csname R@repro/R4/acc\endcsname{98.22}
\expandafter\def\csname R@repro/R6/acc\endcsname{100.00}
\expandafter\def\csname R@rfive/gt-cpu1/1000\endcsname{1.94}
\expandafter\def\csname R@rfive/gt-cpu1/5000\endcsname{1.97}
\expandafter\def\csname R@rfive/gt-cpu4/1000\endcsname{2.36}
\expandafter\def\csname R@rfive/gt-cpu4/5000\endcsname{3.13}
\expandafter\def\csname R@rfive/pytm-cpu1/1000\endcsname{2.04}
\expandafter\def\csname R@rfive/pytm-cpu1/5000\endcsname{2.14}
\expandafter\def\csname R@rfive/pytmpar-cpu4/1000\endcsname{2.19}
\expandafter\def\csname R@rfive/pytmpar-cpu4/5000\endcsname{1.76}
\expandafter\def\csname R@rss/B1/gt-cpu16/default\endcsname{0.1}
\expandafter\def\csname R@rss/B1/pytm-cpu1/default\endcsname{0.1}
\expandafter\def\csname R@rss/B1/pytmcuda-cuda/default\endcsname{0.2}
\expandafter\def\csname R@rss/B1/pytmpar-cpu16/default\endcsname{0.1}
\expandafter\def\csname R@rss/B1/tmu-cpu1/default\endcsname{0.1}
\expandafter\def\csname R@rss/B1/tmu-cuda/default\endcsname{0.3}
\expandafter\def\csname R@rss/B1/tt-cuda/b100\endcsname{1.4}
\expandafter\def\csname R@rss/B1/tt-cuda/batch\endcsname{1.4}
\expandafter\def\csname R@rss/B1/tt-cuda/seq\endcsname{1.4}
\expandafter\def\csname R@rss/B10/pytmcuda-cuda/default\endcsname{0.2}
\expandafter\def\csname R@rss/B10/tmu-cuda/default\endcsname{0.3}
\expandafter\def\csname R@rss/B10/tt-cuda/b100\endcsname{1.3}
\expandafter\def\csname R@rss/B10/tt-cuda/batch\endcsname{1.3}
\expandafter\def\csname R@rss/B10/tt-cuda/seq\endcsname{1.4}
\expandafter\def\csname R@rss/B2/pytm-cpu1/default\endcsname{0.1}
\expandafter\def\csname R@rss/B2/pytmcuda-cuda/default\endcsname{0.2}
\expandafter\def\csname R@rss/B2/pytmpar-cpu16/default\endcsname{0.1}
\expandafter\def\csname R@rss/B2/tmu-cpu1/default\endcsname{0.1}
\expandafter\def\csname R@rss/B2/tmu-cuda/default\endcsname{0.3}
\expandafter\def\csname R@rss/B2/tt-cuda/b100\endcsname{1.1}
\expandafter\def\csname R@rss/B2/tt-cuda/batch\endcsname{1.1}
\expandafter\def\csname R@rss/B2/tt-cuda/seq\endcsname{1.4}
\expandafter\def\csname R@rss/B3/gt-cpu16/default\endcsname{0.1}
\expandafter\def\csname R@rss/B3/pytm-cpu1/default\endcsname{0.1}
\expandafter\def\csname R@rss/B3/pytmcuda-cuda/default\endcsname{0.2}
\expandafter\def\csname R@rss/B3/pytmpar-cpu16/default\endcsname{0.1}
\expandafter\def\csname R@rss/B3/tmu-cpu1/default\endcsname{0.1}
\expandafter\def\csname R@rss/B3/tmu-cuda/default\endcsname{0.3}
\expandafter\def\csname R@rss/B3/tt-cuda/b100\endcsname{1.4}
\expandafter\def\csname R@rss/B3/tt-cuda/batch\endcsname{1.4}
\expandafter\def\csname R@rss/B3/tt-cuda/seq\endcsname{1.4}
\expandafter\def\csname R@rss/B4/gt-cpu16/default\endcsname{0.3}
\expandafter\def\csname R@rss/B4/pytm-cpu1/default\endcsname{0.7}
\expandafter\def\csname R@rss/B4/pytmcuda-cuda/default\endcsname{0.7}
\expandafter\def\csname R@rss/B4/pytmpar-cpu16/default\endcsname{1.3}
\expandafter\def\csname R@rss/B4/tmu-cpu1/default\endcsname{0.6}
\expandafter\def\csname R@rss/B4/tmu-cuda/default\endcsname{0.7}
\expandafter\def\csname R@rss/B4/tt-cpu16/b100\endcsname{2.3}
\expandafter\def\csname R@rss/B4/tt-cpu16/batch\endcsname{2.1}
\expandafter\def\csname R@rss/B4/tt-cuda/b100\endcsname{1.4}
\expandafter\def\csname R@rss/B4/tt-cuda/batch\endcsname{1.4}
\expandafter\def\csname R@rss/B5/gt-cpu16/default\endcsname{0.3}
\expandafter\def\csname R@rss/B5/pytm-cpu1/default\endcsname{0.7}
\expandafter\def\csname R@rss/B5/pytmcuda-cuda/default\endcsname{0.7}
\expandafter\def\csname R@rss/B5/pytmpar-cpu16/default\endcsname{0.9}
\expandafter\def\csname R@rss/B5/tmu-cpu1/default\endcsname{0.6}
\expandafter\def\csname R@rss/B5/tmu-cuda/default\endcsname{0.7}
\expandafter\def\csname R@rss/B5/tt-cpu16/b100\endcsname{2.2}
\expandafter\def\csname R@rss/B5/tt-cpu16/batch\endcsname{2.2}
\expandafter\def\csname R@rss/B5/tt-cuda/b100\endcsname{1.4}
\expandafter\def\csname R@rss/B5/tt-cuda/batch\endcsname{1.4}
\expandafter\def\csname R@rss/B7/pytm-cpu1/default\endcsname{1.2}
\expandafter\def\csname R@rss/B7/pytmcuda-cuda/default\endcsname{0.7}
\expandafter\def\csname R@rss/B7/pytmpar-cpu16/default\endcsname{1.3}
\expandafter\def\csname R@rss/B7/tmu-cpu1/default\endcsname{1.4}
\expandafter\def\csname R@rss/B7/tmu-cuda/default\endcsname{1.4}
\expandafter\def\csname R@rss/B7/tt-cuda/b100\endcsname{1.2}
\expandafter\def\csname R@rss/B7/tt-cuda/batch\endcsname{1.2}
\expandafter\def\csname R@rss/B8/pytmcuda-cuda/default\endcsname{0.7}
\expandafter\def\csname R@rss/B8/tmu-cuda/default\endcsname{1.4}
\expandafter\def\csname R@rss/B8/tt-cuda/b100\endcsname{1.2}
\expandafter\def\csname R@rss/B8/tt-cuda/batch\endcsname{1.2}
\expandafter\def\csname R@rss/B9/pytmcuda-cuda/default\endcsname{2.1}
\expandafter\def\csname R@rss/B9/tmu-cuda/default\endcsname{2.0}
\expandafter\def\csname R@rss/B9/tt-cuda/b100\endcsname{3.6}
\expandafter\def\csname R@rss/B9/tt-cuda/batch\endcsname{3.6}
\expandafter\def\csname R@runs/B1/gt/default\endcsname{1}
\expandafter\def\csname R@runs/B1/pytm/default\endcsname{1}
\expandafter\def\csname R@runs/B1/pytmcuda/default\endcsname{5}
\expandafter\def\csname R@runs/B1/pytmpar/default\endcsname{1}
\expandafter\def\csname R@runs/B1/tmu/default\endcsname{5}
\expandafter\def\csname R@runs/B1/tt/b100\endcsname{5}
\expandafter\def\csname R@runs/B1/tt/batch\endcsname{5}
\expandafter\def\csname R@runs/B1/tt/seq\endcsname{5}
\expandafter\def\csname R@runs/B10/pytmcuda/default\endcsname{25}
\expandafter\def\csname R@runs/B10/tmu/default\endcsname{25}
\expandafter\def\csname R@runs/B10/tt/b100\endcsname{25}
\expandafter\def\csname R@runs/B10/tt/batch\endcsname{25}
\expandafter\def\csname R@runs/B10/tt/seq\endcsname{25}
\expandafter\def\csname R@runs/B2/pytm/default\endcsname{1}
\expandafter\def\csname R@runs/B2/pytmcuda/default\endcsname{5}
\expandafter\def\csname R@runs/B2/pytmpar/default\endcsname{1}
\expandafter\def\csname R@runs/B2/tmu/default\endcsname{5}
\expandafter\def\csname R@runs/B2/tt/b100\endcsname{5}
\expandafter\def\csname R@runs/B2/tt/batch\endcsname{5}
\expandafter\def\csname R@runs/B2/tt/seq\endcsname{5}
\expandafter\def\csname R@runs/B3/gt/default\endcsname{1}
\expandafter\def\csname R@runs/B3/pytm/default\endcsname{1}
\expandafter\def\csname R@runs/B3/pytmcuda/default\endcsname{100}
\expandafter\def\csname R@runs/B3/pytmpar/default\endcsname{1}
\expandafter\def\csname R@runs/B3/tmu/default\endcsname{100}
\expandafter\def\csname R@runs/B3/tt/b100\endcsname{100}
\expandafter\def\csname R@runs/B3/tt/batch\endcsname{100}
\expandafter\def\csname R@runs/B3/tt/seq\endcsname{100}
\expandafter\def\csname R@runs/B4/gt/default\endcsname{1}
\expandafter\def\csname R@runs/B4/pytm/default\endcsname{1}
\expandafter\def\csname R@runs/B4/pytmcuda/default\endcsname{5}
\expandafter\def\csname R@runs/B4/pytmpar/default\endcsname{1}
\expandafter\def\csname R@runs/B4/tmu/default\endcsname{5}
\expandafter\def\csname R@runs/B4/tt/b100\endcsname{5}
\expandafter\def\csname R@runs/B4/tt/batch\endcsname{5}
\expandafter\def\csname R@runs/B5/gt/default\endcsname{1}
\expandafter\def\csname R@runs/B5/pytm/default\endcsname{1}
\expandafter\def\csname R@runs/B5/pytmcuda/default\endcsname{5}
\expandafter\def\csname R@runs/B5/pytmpar/default\endcsname{1}
\expandafter\def\csname R@runs/B5/tmu/default\endcsname{5}
\expandafter\def\csname R@runs/B5/tt/b100\endcsname{5}
\expandafter\def\csname R@runs/B5/tt/batch\endcsname{5}
\expandafter\def\csname R@runs/B7/pytm/default\endcsname{1}
\expandafter\def\csname R@runs/B7/pytmcuda/default\endcsname{5}
\expandafter\def\csname R@runs/B7/pytmpar/default\endcsname{1}
\expandafter\def\csname R@runs/B7/tmu/default\endcsname{5}
\expandafter\def\csname R@runs/B7/tt/b100\endcsname{5}
\expandafter\def\csname R@runs/B7/tt/batch\endcsname{5}
\expandafter\def\csname R@runs/B8/pytmcuda/default\endcsname{5}
\expandafter\def\csname R@runs/B8/tmu/default\endcsname{5}
\expandafter\def\csname R@runs/B8/tt/b100\endcsname{5}
\expandafter\def\csname R@runs/B8/tt/batch\endcsname{5}
\expandafter\def\csname R@runs/B9/pytmcuda/default\endcsname{5}
\expandafter\def\csname R@runs/B9/tmu/default\endcsname{5}
\expandafter\def\csname R@runs/B9/tt/b100\endcsname{5}
\expandafter\def\csname R@runs/B9/tt/batch\endcsname{5}
\expandafter\def\csname R@sd/B1/pytmcuda/default\endcsname{0.00}
\expandafter\def\csname R@sd/B1/tmu/default\endcsname{0.00}
\expandafter\def\csname R@sd/B1/tt/b100\endcsname{0.00}
\expandafter\def\csname R@sd/B1/tt/batch\endcsname{0.00}
\expandafter\def\csname R@sd/B1/tt/seq\endcsname{0.00}
\expandafter\def\csname R@sd/B10/pytmcuda/default\endcsname{0.010}
\expandafter\def\csname R@sd/B10/tmu/default\endcsname{0.015}
\expandafter\def\csname R@sd/B10/tt/b100\endcsname{0.016}
\expandafter\def\csname R@sd/B10/tt/batch\endcsname{0.014}
\expandafter\def\csname R@sd/B10/tt/seq\endcsname{0.011}
\expandafter\def\csname R@sd/B2/pytmcuda/default\endcsname{1.22}
\expandafter\def\csname R@sd/B2/tmu/default\endcsname{0.56}
\expandafter\def\csname R@sd/B2/tt/b100\endcsname{3.83}
\expandafter\def\csname R@sd/B2/tt/batch\endcsname{0.27}
\expandafter\def\csname R@sd/B2/tt/seq\endcsname{0.46}
\expandafter\def\csname R@sd/B3/pytmcuda/default\endcsname{1.30}
\expandafter\def\csname R@sd/B3/tmu/default\endcsname{1.25}
\expandafter\def\csname R@sd/B3/tt/b100\endcsname{1.74}
\expandafter\def\csname R@sd/B3/tt/batch\endcsname{1.48}
\expandafter\def\csname R@sd/B3/tt/seq\endcsname{1.37}
\expandafter\def\csname R@sd/B4/pytmcuda/default\endcsname{0.05}
\expandafter\def\csname R@sd/B4/tmu/default\endcsname{0.15}
\expandafter\def\csname R@sd/B4/tt/b100\endcsname{0.13}
\expandafter\def\csname R@sd/B4/tt/batch\endcsname{0.12}
\expandafter\def\csname R@sd/B5/pytmcuda/default\endcsname{0.08}
\expandafter\def\csname R@sd/B5/tmu/default\endcsname{0.12}
\expandafter\def\csname R@sd/B5/tt/b100\endcsname{0.06}
\expandafter\def\csname R@sd/B5/tt/batch\endcsname{0.08}
\expandafter\def\csname R@sd/B7/pytmcuda/default\endcsname{0.04}
\expandafter\def\csname R@sd/B7/tmu/default\endcsname{0.09}
\expandafter\def\csname R@sd/B7/tt/b100\endcsname{0.07}
\expandafter\def\csname R@sd/B7/tt/batch\endcsname{0.05}
\expandafter\def\csname R@sd/B8/pytmcuda/default\endcsname{2.08}
\expandafter\def\csname R@sd/B8/tmu/default\endcsname{0.17}
\expandafter\def\csname R@sd/B8/tt/b100\endcsname{0.10}
\expandafter\def\csname R@sd/B8/tt/batch\endcsname{0.18}
\expandafter\def\csname R@sd/B9/pytmcuda/default\endcsname{4.45}
\expandafter\def\csname R@sd/B9/tmu/default\endcsname{0.47}
\expandafter\def\csname R@sd/B9/tt/b100\endcsname{0.63}
\expandafter\def\csname R@sd/B9/tt/batch\endcsname{0.93}
\expandafter\def\csname R@statemib/B1/gt-cpu16/default\endcsname{0.0}
\expandafter\def\csname R@statemib/B1/pytm-cpu1/default\endcsname{0.0}
\expandafter\def\csname R@statemib/B1/pytmcuda-cuda/default\endcsname{0.0}
\expandafter\def\csname R@statemib/B1/pytmpar-cpu16/default\endcsname{0.0}
\expandafter\def\csname R@statemib/B1/tmu-cpu1/default\endcsname{0.0}
\expandafter\def\csname R@statemib/B1/tmu-cuda/default\endcsname{0.0}
\expandafter\def\csname R@statemib/B1/tt-cuda/b100\endcsname{0.0}
\expandafter\def\csname R@statemib/B1/tt-cuda/batch\endcsname{0.0}
\expandafter\def\csname R@statemib/B1/tt-cuda/seq\endcsname{0.0}
\expandafter\def\csname R@statemib/B10/pytmcuda-cuda/default\endcsname{0.2}
\expandafter\def\csname R@statemib/B10/tmu-cuda/default\endcsname{0.2}
\expandafter\def\csname R@statemib/B10/tt-cuda/b100\endcsname{0.2}
\expandafter\def\csname R@statemib/B10/tt-cuda/batch\endcsname{0.2}
\expandafter\def\csname R@statemib/B10/tt-cuda/seq\endcsname{0.2}
\expandafter\def\csname R@statemib/B2/pytm-cpu1/default\endcsname{0.0}
\expandafter\def\csname R@statemib/B2/pytmcuda-cuda/default\endcsname{0.0}
\expandafter\def\csname R@statemib/B2/pytmpar-cpu16/default\endcsname{0.0}
\expandafter\def\csname R@statemib/B2/tmu-cpu1/default\endcsname{0.0}
\expandafter\def\csname R@statemib/B2/tmu-cuda/default\endcsname{0.0}
\expandafter\def\csname R@statemib/B2/tt-cuda/b100\endcsname{0.0}
\expandafter\def\csname R@statemib/B2/tt-cuda/batch\endcsname{0.0}
\expandafter\def\csname R@statemib/B2/tt-cuda/seq\endcsname{0.0}
\expandafter\def\csname R@statemib/B3/gt-cpu16/default\endcsname{0.9}
\expandafter\def\csname R@statemib/B3/pytm-cpu1/default\endcsname{0.9}
\expandafter\def\csname R@statemib/B3/pytmcuda-cuda/default\endcsname{0.9}
\expandafter\def\csname R@statemib/B3/pytmpar-cpu16/default\endcsname{0.9}
\expandafter\def\csname R@statemib/B3/tmu-cpu1/default\endcsname{1.9}
\expandafter\def\csname R@statemib/B3/tmu-cuda/default\endcsname{0.9}
\expandafter\def\csname R@statemib/B3/tt-cuda/b100\endcsname{0.9}
\expandafter\def\csname R@statemib/B3/tt-cuda/batch\endcsname{0.9}
\expandafter\def\csname R@statemib/B3/tt-cuda/seq\endcsname{0.9}
\expandafter\def\csname R@statemib/B4/gt-cpu16/default\endcsname{30.3}
\expandafter\def\csname R@statemib/B4/pytm-cpu1/default\endcsname{30.0}
\expandafter\def\csname R@statemib/B4/pytmcuda-cuda/default\endcsname{29.9}
\expandafter\def\csname R@statemib/B4/pytmpar-cpu16/default\endcsname{30.0}
\expandafter\def\csname R@statemib/B4/tmu-cpu1/default\endcsname{59.9}
\expandafter\def\csname R@statemib/B4/tmu-cuda/default\endcsname{30.0}
\expandafter\def\csname R@statemib/B4/tt-cpu16/b100\endcsname{30.0}
\expandafter\def\csname R@statemib/B4/tt-cpu16/batch\endcsname{30.0}
\expandafter\def\csname R@statemib/B4/tt-cuda/b100\endcsname{30.0}
\expandafter\def\csname R@statemib/B4/tt-cuda/batch\endcsname{30.0}
\expandafter\def\csname R@statemib/B5/gt-cpu16/default\endcsname{30.3}
\expandafter\def\csname R@statemib/B5/pytm-cpu1/default\endcsname{30.0}
\expandafter\def\csname R@statemib/B5/pytmcuda-cuda/default\endcsname{29.9}
\expandafter\def\csname R@statemib/B5/pytmpar-cpu16/default\endcsname{30.0}
\expandafter\def\csname R@statemib/B5/tmu-cpu1/default\endcsname{59.9}
\expandafter\def\csname R@statemib/B5/tmu-cuda/default\endcsname{30.0}
\expandafter\def\csname R@statemib/B5/tt-cpu16/b100\endcsname{30.0}
\expandafter\def\csname R@statemib/B5/tt-cpu16/batch\endcsname{30.0}
\expandafter\def\csname R@statemib/B5/tt-cuda/b100\endcsname{30.0}
\expandafter\def\csname R@statemib/B5/tt-cuda/batch\endcsname{30.0}
\expandafter\def\csname R@statemib/B7/pytm-cpu1/default\endcsname{5.6}
\expandafter\def\csname R@statemib/B7/pytmcuda-cuda/default\endcsname{5.5}
\expandafter\def\csname R@statemib/B7/pytmpar-cpu16/default\endcsname{5.6}
\expandafter\def\csname R@statemib/B7/tmu-cpu1/default\endcsname{11.1}
\expandafter\def\csname R@statemib/B7/tmu-cuda/default\endcsname{5.6}
\expandafter\def\csname R@statemib/B7/tt-cuda/b100\endcsname{5.3}
\expandafter\def\csname R@statemib/B7/tt-cuda/batch\endcsname{5.3}
\expandafter\def\csname R@statemib/B8/pytmcuda-cuda/default\endcsname{5.5}
\expandafter\def\csname R@statemib/B8/tmu-cuda/default\endcsname{5.6}
\expandafter\def\csname R@statemib/B8/tt-cuda/b100\endcsname{5.3}
\expandafter\def\csname R@statemib/B8/tt-cuda/batch\endcsname{5.3}
\expandafter\def\csname R@statemib/B9/pytmcuda-cuda/default\endcsname{191.1}
\expandafter\def\csname R@statemib/B9/tmu-cuda/default\endcsname{191.2}
\expandafter\def\csname R@statemib/B9/tt-cuda/b100\endcsname{190.8}
\expandafter\def\csname R@statemib/B9/tt-cuda/batch\endcsname{190.8}
\expandafter\def\csname R@sweep/100\endcsname{92.5}
\expandafter\def\csname R@sweep/16\endcsname{99.6}
\expandafter\def\csname R@sweep/200\endcsname{75.8}
\expandafter\def\csname R@sweep/32\endcsname{99.7}
\expandafter\def\csname R@sweep/50\endcsname{99.6}
\expandafter\def\csname R@sweep/500\endcsname{60.5}
\expandafter\def\csname R@sweepsd/100\endcsname{3.8}
\expandafter\def\csname R@sweepsd/16\endcsname{0.3}
\expandafter\def\csname R@sweepsd/200\endcsname{4.0}
\expandafter\def\csname R@sweepsd/32\endcsname{0.3}
\expandafter\def\csname R@sweepsd/50\endcsname{0.5}
\expandafter\def\csname R@sweepsd/500\endcsname{9.7}
\expandafter\def\csname R@test/B1/gt-cpu16/default\endcsname{0.204}
\expandafter\def\csname R@test/B1/pytm-cpu1/default\endcsname{0.00264}
\expandafter\def\csname R@test/B1/pytmcuda-cuda/default\endcsname{0.000570}
\expandafter\def\csname R@test/B1/pytmpar-cpu16/default\endcsname{0.00112}
\expandafter\def\csname R@test/B1/tmu-cpu1/default\endcsname{0.0871}
\expandafter\def\csname R@test/B1/tmu-cuda/default\endcsname{0.233}
\expandafter\def\csname R@test/B1/tt-cuda/b100\endcsname{0.000660}
\expandafter\def\csname R@test/B1/tt-cuda/batch\endcsname{0.000752}
\expandafter\def\csname R@test/B1/tt-cuda/seq\endcsname{0.000991}
\expandafter\def\csname R@test/B10/pytmcuda-cuda/default\endcsname{0.00220}
\expandafter\def\csname R@test/B10/tmu-cuda/default\endcsname{0.161}
\expandafter\def\csname R@test/B10/tt-cuda/b100\endcsname{0.000728}
\expandafter\def\csname R@test/B10/tt-cuda/batch\endcsname{0.000746}
\expandafter\def\csname R@test/B10/tt-cuda/seq\endcsname{0.000929}
\expandafter\def\csname R@test/B2/pytm-cpu1/default\endcsname{0.0211}
\expandafter\def\csname R@test/B2/pytmcuda-cuda/default\endcsname{0.00499}
\expandafter\def\csname R@test/B2/pytmpar-cpu16/default\endcsname{0.00468}
\expandafter\def\csname R@test/B2/tmu-cpu1/default\endcsname{0.205}
\expandafter\def\csname R@test/B2/tmu-cuda/default\endcsname{0.488}
\expandafter\def\csname R@test/B2/tt-cuda/b100\endcsname{0.0441}
\expandafter\def\csname R@test/B2/tt-cuda/batch\endcsname{0.0444}
\expandafter\def\csname R@test/B2/tt-cuda/seq\endcsname{0.0443}
\expandafter\def\csname R@test/B3/gt-cpu16/default\endcsname{0.00554}
\expandafter\def\csname R@test/B3/pytm-cpu1/default\endcsname{0.00697}
\expandafter\def\csname R@test/B3/pytmcuda-cuda/default\endcsname{0.000483}
\expandafter\def\csname R@test/B3/pytmpar-cpu16/default\endcsname{0.00316}
\expandafter\def\csname R@test/B3/tmu-cpu1/default\endcsname{0.00451}
\expandafter\def\csname R@test/B3/tmu-cuda/default\endcsname{0.00666}
\expandafter\def\csname R@test/B3/tt-cuda/b100\endcsname{0.000352}
\expandafter\def\csname R@test/B3/tt-cuda/batch\endcsname{0.000372}
\expandafter\def\csname R@test/B3/tt-cuda/seq\endcsname{0.000375}
\expandafter\def\csname R@test/B4/gt-cpu16/default\endcsname{6.51}
\expandafter\def\csname R@test/B4/pytm-cpu1/default\endcsname{3.42}
\expandafter\def\csname R@test/B4/pytmcuda-cuda/default\endcsname{0.523}
\expandafter\def\csname R@test/B4/pytmpar-cpu16/default\endcsname{2.41}
\expandafter\def\csname R@test/B4/tmu-cpu1/default\endcsname{2.35}
\expandafter\def\csname R@test/B4/tmu-cuda/default\endcsname{4.62}
\expandafter\def\csname R@test/B4/tt-cpu16/b100\endcsname{1.63}
\expandafter\def\csname R@test/B4/tt-cpu16/batch\endcsname{1.59}
\expandafter\def\csname R@test/B4/tt-cuda/b100\endcsname{0.0167}
\expandafter\def\csname R@test/B4/tt-cuda/batch\endcsname{0.0163}
\expandafter\def\csname R@test/B5/gt-cpu16/default\endcsname{4.31}
\expandafter\def\csname R@test/B5/pytm-cpu1/default\endcsname{3.28}
\expandafter\def\csname R@test/B5/pytmcuda-cuda/default\endcsname{0.550}
\expandafter\def\csname R@test/B5/pytmpar-cpu16/default\endcsname{2.45}
\expandafter\def\csname R@test/B5/tmu-cpu1/default\endcsname{2.40}
\expandafter\def\csname R@test/B5/tmu-cuda/default\endcsname{4.74}
\expandafter\def\csname R@test/B5/tt-cpu16/b100\endcsname{1.59}
\expandafter\def\csname R@test/B5/tt-cpu16/batch\endcsname{1.66}
\expandafter\def\csname R@test/B5/tt-cuda/b100\endcsname{0.0167}
\expandafter\def\csname R@test/B5/tt-cuda/batch\endcsname{0.0164}
\expandafter\def\csname R@test/B7/pytm-cpu1/default\endcsname{289}
\expandafter\def\csname R@test/B7/pytmcuda-cuda/default\endcsname{1.47}
\expandafter\def\csname R@test/B7/pytmpar-cpu16/default\endcsname{20.9}
\expandafter\def\csname R@test/B7/tmu-cpu1/default\endcsname{227}
\expandafter\def\csname R@test/B7/tmu-cuda/default\endcsname{22.0}
\expandafter\def\csname R@test/B7/tt-cuda/b100\endcsname{0.340}
\expandafter\def\csname R@test/B7/tt-cuda/batch\endcsname{0.337}
\expandafter\def\csname R@test/B8/pytmcuda-cuda/default\endcsname{1.47}
\expandafter\def\csname R@test/B8/tmu-cuda/default\endcsname{23.2}
\expandafter\def\csname R@test/B8/tt-cuda/b100\endcsname{0.337}
\expandafter\def\csname R@test/B8/tt-cuda/batch\endcsname{0.336}
\expandafter\def\csname R@test/B9/pytmcuda-cuda/default\endcsname{11.7}
\expandafter\def\csname R@test/B9/tmu-cuda/default\endcsname{15.4}
\expandafter\def\csname R@test/B9/tt-cuda/b100\endcsname{0.168}
\expandafter\def\csname R@test/B9/tt-cuda/batch\endcsname{0.167}
\expandafter\def\csname R@train/B1/gt-cpu16/default\endcsname{0.417}
\expandafter\def\csname R@train/B1/pytm-cpu1/default\endcsname{0.00786}
\expandafter\def\csname R@train/B1/pytmcuda-cuda/default\endcsname{0.0136}
\expandafter\def\csname R@train/B1/pytmpar-cpu16/default\endcsname{0.00657}
\expandafter\def\csname R@train/B1/tmu-cpu1/default\endcsname{0.445}
\expandafter\def\csname R@train/B1/tmu-cuda/default\endcsname{1.13}
\expandafter\def\csname R@train/B1/tt-cuda/b100\endcsname{0.0136}
\expandafter\def\csname R@train/B1/tt-cuda/batch\endcsname{0.0421}
\expandafter\def\csname R@train/B1/tt-cuda/seq\endcsname{0.112}
\expandafter\def\csname R@train/B10/pytmcuda-cuda/default\endcsname{0.124}
\expandafter\def\csname R@train/B10/tmu-cuda/default\endcsname{1.45}
\expandafter\def\csname R@train/B10/tt-cuda/b100\endcsname{0.0311}
\expandafter\def\csname R@train/B10/tt-cuda/batch\endcsname{0.0973}
\expandafter\def\csname R@train/B10/tt-cuda/seq\endcsname{0.429}
\expandafter\def\csname R@train/B2/pytm-cpu1/default\endcsname{0.0118}
\expandafter\def\csname R@train/B2/pytmcuda-cuda/default\endcsname{0.00884}
\expandafter\def\csname R@train/B2/pytmpar-cpu16/default\endcsname{0.00741}
\expandafter\def\csname R@train/B2/tmu-cpu1/default\endcsname{0.226}
\expandafter\def\csname R@train/B2/tmu-cuda/default\endcsname{0.579}
\expandafter\def\csname R@train/B2/tt-cuda/b100\endcsname{0.00902}
\expandafter\def\csname R@train/B2/tt-cuda/batch\endcsname{0.0439}
\expandafter\def\csname R@train/B2/tt-cuda/seq\endcsname{0.102}
\expandafter\def\csname R@train/B3/gt-cpu16/default\endcsname{0.0434}
\expandafter\def\csname R@train/B3/pytm-cpu1/default\endcsname{0.0337}
\expandafter\def\csname R@train/B3/pytmcuda-cuda/default\endcsname{0.00635}
\expandafter\def\csname R@train/B3/pytmpar-cpu16/default\endcsname{0.00643}
\expandafter\def\csname R@train/B3/tmu-cpu1/default\endcsname{0.0766}
\expandafter\def\csname R@train/B3/tmu-cuda/default\endcsname{0.125}
\expandafter\def\csname R@train/B3/tt-cuda/b100\endcsname{0.00153}
\expandafter\def\csname R@train/B3/tt-cuda/batch\endcsname{0.00426}
\expandafter\def\csname R@train/B3/tt-cuda/seq\endcsname{0.0163}
\expandafter\def\csname R@train/B4/gt-cpu16/default\endcsname{46.3}
\expandafter\def\csname R@train/B4/pytm-cpu1/default\endcsname{9.32}
\expandafter\def\csname R@train/B4/pytmcuda-cuda/default\endcsname{3.44}
\expandafter\def\csname R@train/B4/pytmpar-cpu16/default\endcsname{2.99}
\expandafter\def\csname R@train/B4/tmu-cpu1/default\endcsname{32.4}
\expandafter\def\csname R@train/B4/tmu-cuda/default\endcsname{19.3}
\expandafter\def\csname R@train/B4/tt-cpu16/b100\endcsname{64.3}
\expandafter\def\csname R@train/B4/tt-cpu16/batch\endcsname{69.6}
\expandafter\def\csname R@train/B4/tt-cuda/b100\endcsname{0.788}
\expandafter\def\csname R@train/B4/tt-cuda/batch\endcsname{1.59}
\expandafter\def\csname R@train/B5/gt-cpu16/default\endcsname{75.5}
\expandafter\def\csname R@train/B5/pytm-cpu1/default\endcsname{7.44}
\expandafter\def\csname R@train/B5/pytmcuda-cuda/default\endcsname{3.62}
\expandafter\def\csname R@train/B5/pytmpar-cpu16/default\endcsname{3.34}
\expandafter\def\csname R@train/B5/tmu-cpu1/default\endcsname{32.8}
\expandafter\def\csname R@train/B5/tmu-cuda/default\endcsname{20.8}
\expandafter\def\csname R@train/B5/tt-cpu16/b100\endcsname{67.5}
\expandafter\def\csname R@train/B5/tt-cpu16/batch\endcsname{73.0}
\expandafter\def\csname R@train/B5/tt-cuda/b100\endcsname{0.802}
\expandafter\def\csname R@train/B5/tt-cuda/batch\endcsname{1.63}
\expandafter\def\csname R@train/B7/pytm-cpu1/default\endcsname{354}
\expandafter\def\csname R@train/B7/pytmcuda-cuda/default\endcsname{9.53}
\expandafter\def\csname R@train/B7/pytmpar-cpu16/default\endcsname{30.4}
\expandafter\def\csname R@train/B7/tmu-cpu1/default\endcsname{276}
\expandafter\def\csname R@train/B7/tmu-cuda/default\endcsname{41.2}
\expandafter\def\csname R@train/B7/tt-cuda/b100\endcsname{3.99}
\expandafter\def\csname R@train/B7/tt-cuda/batch\endcsname{4.65}
\expandafter\def\csname R@train/B8/pytmcuda-cuda/default\endcsname{9.58}
\expandafter\def\csname R@train/B8/tmu-cuda/default\endcsname{47.8}
\expandafter\def\csname R@train/B8/tt-cuda/b100\endcsname{3.97}
\expandafter\def\csname R@train/B8/tt-cuda/batch\endcsname{4.66}
\expandafter\def\csname R@train/B9/pytmcuda-cuda/default\endcsname{21.2}
\expandafter\def\csname R@train/B9/tmu-cuda/default\endcsname{50.9}
\expandafter\def\csname R@train/B9/tt-cuda/b100\endcsname{1.94}
\expandafter\def\csname R@train/B9/tt-cuda/batch\endcsname{4.07}
\expandafter\def\csname R@version/gt\endcsname{1.1.0}
\expandafter\def\csname R@version/pytm\endcsname{0.6.6}
\expandafter\def\csname R@version/pytmcuda\endcsname{0.1.9}
\expandafter\def\csname R@version/pytmpar\endcsname{0.2.1}
\expandafter\def\csname R@version/tmu\endcsname{0.8.3}
\expandafter\def\csname R@version/tt\endcsname{0.6.0, 0.6.1}
\expandafter\def\csname R@xbsix/gt-cpu1\endcsname{59}
\expandafter\def\csname R@xbsix/gt-cpu16\endcsname{26}
\expandafter\def\csname R@xbsix/gt-cpu4\endcsname{32}
\expandafter\def\csname R@xbsix/pytm-cpu1\endcsname{141}
\expandafter\def\csname R@xbsix/pytmcuda-cuda\endcsname{30}
\expandafter\def\csname R@xbsix/pytmpar-cpu16\endcsname{38}
\expandafter\def\csname R@xbsix/pytmpar-cpu4\endcsname{88}
\expandafter\def\csname R@xbsix/tmu-cpu1\endcsname{306}
\expandafter\def\csname R@xbsix/tmu-cuda\endcsname{62}
\expandafter\def\csname R@xbsix/tt-cuda[b100]\endcsname{1}
\expandafter\def\csname R@xbsix/tt-cuda[batch]\endcsname{2}
\expandafter\def\csname R@xtest/B1/gt-cpu16/b100\endcsname{309.6}
\expandafter\def\csname R@xtest/B1/gt-cpu16/batch\endcsname{271.9}
\expandafter\def\csname R@xtest/B1/gt-cpu16/seq\endcsname{206.3}
\expandafter\def\csname R@xtest/B1/pytm-cpu1/b100\endcsname{4.0}
\expandafter\def\csname R@xtest/B1/pytm-cpu1/batch\endcsname{3.5}
\expandafter\def\csname R@xtest/B1/pytm-cpu1/seq\endcsname{2.7}
\expandafter\def\csname R@xtest/B1/pytmcuda-cuda/b100\endcsname{0.9}
\expandafter\def\csname R@xtest/B1/pytmcuda-cuda/batch\endcsname{0.8}
\expandafter\def\csname R@xtest/B1/pytmcuda-cuda/seq\endcsname{0.6}
\expandafter\def\csname R@xtest/B1/pytmpar-cpu16/b100\endcsname{1.7}
\expandafter\def\csname R@xtest/B1/pytmpar-cpu16/batch\endcsname{1.5}
\expandafter\def\csname R@xtest/B1/pytmpar-cpu16/seq\endcsname{1.1}
\expandafter\def\csname R@xtest/B1/tmu-cpu1/b100\endcsname{131.8}
\expandafter\def\csname R@xtest/B1/tmu-cpu1/batch\endcsname{115.8}
\expandafter\def\csname R@xtest/B1/tmu-cpu1/seq\endcsname{87.8}
\expandafter\def\csname R@xtest/B1/tmu-cuda/b100\endcsname{353.1}
\expandafter\def\csname R@xtest/B1/tmu-cuda/batch\endcsname{310.0}
\expandafter\def\csname R@xtest/B1/tmu-cuda/seq\endcsname{235.2}
\expandafter\def\csname R@xtest/B10/pytmcuda-cuda/b100\endcsname{3.0}
\expandafter\def\csname R@xtest/B10/pytmcuda-cuda/batch\endcsname{3.0}
\expandafter\def\csname R@xtest/B10/pytmcuda-cuda/seq\endcsname{2.4}
\expandafter\def\csname R@xtest/B10/tmu-cuda/b100\endcsname{221.4}
\expandafter\def\csname R@xtest/B10/tmu-cuda/batch\endcsname{216.2}
\expandafter\def\csname R@xtest/B10/tmu-cuda/seq\endcsname{173.5}
\expandafter\def\csname R@xtest/B2/pytm-cpu1/b100\endcsname{0.5}
\expandafter\def\csname R@xtest/B2/pytm-cpu1/batch\endcsname{0.5}
\expandafter\def\csname R@xtest/B2/pytm-cpu1/seq\endcsname{0.5}
\expandafter\def\csname R@xtest/B2/pytmcuda-cuda/b100\endcsname{0.1}
\expandafter\def\csname R@xtest/B2/pytmcuda-cuda/batch\endcsname{0.1}
\expandafter\def\csname R@xtest/B2/pytmcuda-cuda/seq\endcsname{0.1}
\expandafter\def\csname R@xtest/B2/pytmpar-cpu16/b100\endcsname{0.1}
\expandafter\def\csname R@xtest/B2/pytmpar-cpu16/batch\endcsname{0.1}
\expandafter\def\csname R@xtest/B2/pytmpar-cpu16/seq\endcsname{0.1}
\expandafter\def\csname R@xtest/B2/tmu-cpu1/b100\endcsname{4.6}
\expandafter\def\csname R@xtest/B2/tmu-cpu1/batch\endcsname{4.6}
\expandafter\def\csname R@xtest/B2/tmu-cpu1/seq\endcsname{4.6}
\expandafter\def\csname R@xtest/B2/tmu-cuda/b100\endcsname{11.1}
\expandafter\def\csname R@xtest/B2/tmu-cuda/batch\endcsname{11.0}
\expandafter\def\csname R@xtest/B2/tmu-cuda/seq\endcsname{11.0}
\expandafter\def\csname R@xtest/B3/gt-cpu16/b100\endcsname{15.7}
\expandafter\def\csname R@xtest/B3/gt-cpu16/batch\endcsname{14.9}
\expandafter\def\csname R@xtest/B3/gt-cpu16/seq\endcsname{14.8}
\expandafter\def\csname R@xtest/B3/pytm-cpu1/b100\endcsname{19.8}
\expandafter\def\csname R@xtest/B3/pytm-cpu1/batch\endcsname{18.8}
\expandafter\def\csname R@xtest/B3/pytm-cpu1/seq\endcsname{18.6}
\expandafter\def\csname R@xtest/B3/pytmcuda-cuda/b100\endcsname{1.4}
\expandafter\def\csname R@xtest/B3/pytmcuda-cuda/batch\endcsname{1.3}
\expandafter\def\csname R@xtest/B3/pytmcuda-cuda/seq\endcsname{1.3}
\expandafter\def\csname R@xtest/B3/pytmpar-cpu16/b100\endcsname{9.0}
\expandafter\def\csname R@xtest/B3/pytmpar-cpu16/batch\endcsname{8.5}
\expandafter\def\csname R@xtest/B3/pytmpar-cpu16/seq\endcsname{8.4}
\expandafter\def\csname R@xtest/B3/tmu-cpu1/b100\endcsname{12.8}
\expandafter\def\csname R@xtest/B3/tmu-cpu1/batch\endcsname{12.1}
\expandafter\def\csname R@xtest/B3/tmu-cpu1/seq\endcsname{12.0}
\expandafter\def\csname R@xtest/B3/tmu-cuda/b100\endcsname{18.9}
\expandafter\def\csname R@xtest/B3/tmu-cuda/batch\endcsname{17.9}
\expandafter\def\csname R@xtest/B3/tmu-cuda/seq\endcsname{17.8}
\expandafter\def\csname R@xtest/B4/gt-cpu16/b100\endcsname{390.1}
\expandafter\def\csname R@xtest/B4/gt-cpu16/batch\endcsname{398.9}
\expandafter\def\csname R@xtest/B4/pytm-cpu1/b100\endcsname{204.8}
\expandafter\def\csname R@xtest/B4/pytm-cpu1/batch\endcsname{209.4}
\expandafter\def\csname R@xtest/B4/pytmcuda-cuda/b100\endcsname{31.4}
\expandafter\def\csname R@xtest/B4/pytmcuda-cuda/batch\endcsname{32.1}
\expandafter\def\csname R@xtest/B4/pytmpar-cpu16/b100\endcsname{144.2}
\expandafter\def\csname R@xtest/B4/pytmpar-cpu16/batch\endcsname{147.5}
\expandafter\def\csname R@xtest/B4/tmu-cpu1/b100\endcsname{141.0}
\expandafter\def\csname R@xtest/B4/tmu-cpu1/batch\endcsname{144.2}
\expandafter\def\csname R@xtest/B4/tmu-cuda/b100\endcsname{277.1}
\expandafter\def\csname R@xtest/B4/tmu-cuda/batch\endcsname{283.4}
\expandafter\def\csname R@xtest/B5/gt-cpu16/b100\endcsname{259.0}
\expandafter\def\csname R@xtest/B5/gt-cpu16/batch\endcsname{263.4}
\expandafter\def\csname R@xtest/B5/pytm-cpu1/b100\endcsname{196.8}
\expandafter\def\csname R@xtest/B5/pytm-cpu1/batch\endcsname{200.1}
\expandafter\def\csname R@xtest/B5/pytmcuda-cuda/b100\endcsname{33.0}
\expandafter\def\csname R@xtest/B5/pytmcuda-cuda/batch\endcsname{33.6}
\expandafter\def\csname R@xtest/B5/pytmpar-cpu16/b100\endcsname{147.0}
\expandafter\def\csname R@xtest/B5/pytmpar-cpu16/batch\endcsname{149.4}
\expandafter\def\csname R@xtest/B5/tmu-cpu1/b100\endcsname{143.9}
\expandafter\def\csname R@xtest/B5/tmu-cpu1/batch\endcsname{146.3}
\expandafter\def\csname R@xtest/B5/tmu-cuda/b100\endcsname{284.7}
\expandafter\def\csname R@xtest/B5/tmu-cuda/batch\endcsname{289.5}
\expandafter\def\csname R@xtest/B7/pytm-cpu1/b100\endcsname{849.1}
\expandafter\def\csname R@xtest/B7/pytm-cpu1/batch\endcsname{858.2}
\expandafter\def\csname R@xtest/B7/pytmcuda-cuda/b100\endcsname{4.3}
\expandafter\def\csname R@xtest/B7/pytmcuda-cuda/batch\endcsname{4.4}
\expandafter\def\csname R@xtest/B7/pytmpar-cpu16/b100\endcsname{61.5}
\expandafter\def\csname R@xtest/B7/pytmpar-cpu16/batch\endcsname{62.2}
\expandafter\def\csname R@xtest/B7/tmu-cpu1/b100\endcsname{666.8}
\expandafter\def\csname R@xtest/B7/tmu-cpu1/batch\endcsname{673.9}
\expandafter\def\csname R@xtest/B7/tmu-cuda/b100\endcsname{64.6}
\expandafter\def\csname R@xtest/B7/tmu-cuda/batch\endcsname{65.3}
\expandafter\def\csname R@xtest/B8/pytmcuda-cuda/b100\endcsname{4.4}
\expandafter\def\csname R@xtest/B8/pytmcuda-cuda/batch\endcsname{4.4}
\expandafter\def\csname R@xtest/B8/tmu-cuda/b100\endcsname{68.6}
\expandafter\def\csname R@xtest/B8/tmu-cuda/batch\endcsname{68.8}
\expandafter\def\csname R@xtest/B9/pytmcuda-cuda/b100\endcsname{69.5}
\expandafter\def\csname R@xtest/B9/pytmcuda-cuda/batch\endcsname{70.0}
\expandafter\def\csname R@xtest/B9/tmu-cuda/b100\endcsname{91.3}
\expandafter\def\csname R@xtest/B9/tmu-cuda/batch\endcsname{92.0}
\expandafter\def\csname R@xtrain/B1/gt-cpu16/b100\endcsname{30.6}
\expandafter\def\csname R@xtrain/B1/gt-cpu16/batch\endcsname{9.9}
\expandafter\def\csname R@xtrain/B1/gt-cpu16/seq\endcsname{3.7}
\expandafter\def\csname R@xtrain/B1/pytm-cpu1/b100\endcsname{0.6}
\expandafter\def\csname R@xtrain/B1/pytm-cpu1/batch\endcsname{0.2}
\expandafter\def\csname R@xtrain/B1/pytm-cpu1/seq\endcsname{0.1}
\expandafter\def\csname R@xtrain/B1/pytmcuda-cuda/b100\endcsname{1.0}
\expandafter\def\csname R@xtrain/B1/pytmcuda-cuda/batch\endcsname{0.3}
\expandafter\def\csname R@xtrain/B1/pytmcuda-cuda/seq\endcsname{0.1}
\expandafter\def\csname R@xtrain/B1/pytmpar-cpu16/b100\endcsname{0.5}
\expandafter\def\csname R@xtrain/B1/pytmpar-cpu16/batch\endcsname{0.2}
\expandafter\def\csname R@xtrain/B1/pytmpar-cpu16/seq\endcsname{0.1}
\expandafter\def\csname R@xtrain/B1/tmu-cpu1/b100\endcsname{32.7}
\expandafter\def\csname R@xtrain/B1/tmu-cpu1/batch\endcsname{10.6}
\expandafter\def\csname R@xtrain/B1/tmu-cpu1/seq\endcsname{4.0}
\expandafter\def\csname R@xtrain/B1/tmu-cuda/b100\endcsname{83.4}
\expandafter\def\csname R@xtrain/B1/tmu-cuda/batch\endcsname{26.9}
\expandafter\def\csname R@xtrain/B1/tmu-cuda/seq\endcsname{10.1}
\expandafter\def\csname R@xtrain/B10/pytmcuda-cuda/b100\endcsname{4.0}
\expandafter\def\csname R@xtrain/B10/pytmcuda-cuda/batch\endcsname{1.3}
\expandafter\def\csname R@xtrain/B10/pytmcuda-cuda/seq\endcsname{0.3}
\expandafter\def\csname R@xtrain/B10/tmu-cuda/b100\endcsname{46.5}
\expandafter\def\csname R@xtrain/B10/tmu-cuda/batch\endcsname{14.9}
\expandafter\def\csname R@xtrain/B10/tmu-cuda/seq\endcsname{3.4}
\expandafter\def\csname R@xtrain/B2/pytm-cpu1/b100\endcsname{1.3}
\expandafter\def\csname R@xtrain/B2/pytm-cpu1/batch\endcsname{0.3}
\expandafter\def\csname R@xtrain/B2/pytm-cpu1/seq\endcsname{0.1}
\expandafter\def\csname R@xtrain/B2/pytmcuda-cuda/b100\endcsname{1.0}
\expandafter\def\csname R@xtrain/B2/pytmcuda-cuda/batch\endcsname{0.2}
\expandafter\def\csname R@xtrain/B2/pytmcuda-cuda/seq\endcsname{0.1}
\expandafter\def\csname R@xtrain/B2/pytmpar-cpu16/b100\endcsname{0.8}
\expandafter\def\csname R@xtrain/B2/pytmpar-cpu16/batch\endcsname{0.2}
\expandafter\def\csname R@xtrain/B2/pytmpar-cpu16/seq\endcsname{0.1}
\expandafter\def\csname R@xtrain/B2/tmu-cpu1/b100\endcsname{25.0}
\expandafter\def\csname R@xtrain/B2/tmu-cpu1/batch\endcsname{5.1}
\expandafter\def\csname R@xtrain/B2/tmu-cpu1/seq\endcsname{2.2}
\expandafter\def\csname R@xtrain/B2/tmu-cuda/b100\endcsname{64.2}
\expandafter\def\csname R@xtrain/B2/tmu-cuda/batch\endcsname{13.2}
\expandafter\def\csname R@xtrain/B2/tmu-cuda/seq\endcsname{5.7}
\expandafter\def\csname R@xtrain/B3/gt-cpu16/b100\endcsname{28.3}
\expandafter\def\csname R@xtrain/B3/gt-cpu16/batch\endcsname{10.2}
\expandafter\def\csname R@xtrain/B3/gt-cpu16/seq\endcsname{2.7}
\expandafter\def\csname R@xtrain/B3/pytm-cpu1/b100\endcsname{22.0}
\expandafter\def\csname R@xtrain/B3/pytm-cpu1/batch\endcsname{7.9}
\expandafter\def\csname R@xtrain/B3/pytm-cpu1/seq\endcsname{2.1}
\expandafter\def\csname R@xtrain/B3/pytmcuda-cuda/b100\endcsname{4.1}
\expandafter\def\csname R@xtrain/B3/pytmcuda-cuda/batch\endcsname{1.5}
\expandafter\def\csname R@xtrain/B3/pytmcuda-cuda/seq\endcsname{0.4}
\expandafter\def\csname R@xtrain/B3/pytmpar-cpu16/b100\endcsname{4.2}
\expandafter\def\csname R@xtrain/B3/pytmpar-cpu16/batch\endcsname{1.5}
\expandafter\def\csname R@xtrain/B3/pytmpar-cpu16/seq\endcsname{0.4}
\expandafter\def\csname R@xtrain/B3/tmu-cpu1/b100\endcsname{49.9}
\expandafter\def\csname R@xtrain/B3/tmu-cpu1/batch\endcsname{18.0}
\expandafter\def\csname R@xtrain/B3/tmu-cpu1/seq\endcsname{4.7}
\expandafter\def\csname R@xtrain/B3/tmu-cuda/b100\endcsname{81.2}
\expandafter\def\csname R@xtrain/B3/tmu-cuda/batch\endcsname{29.2}
\expandafter\def\csname R@xtrain/B3/tmu-cuda/seq\endcsname{7.6}
\expandafter\def\csname R@xtrain/B4/gt-cpu16/b100\endcsname{58.8}
\expandafter\def\csname R@xtrain/B4/gt-cpu16/batch\endcsname{29.1}
\expandafter\def\csname R@xtrain/B4/pytm-cpu1/b100\endcsname{11.8}
\expandafter\def\csname R@xtrain/B4/pytm-cpu1/batch\endcsname{5.9}
\expandafter\def\csname R@xtrain/B4/pytmcuda-cuda/b100\endcsname{4.4}
\expandafter\def\csname R@xtrain/B4/pytmcuda-cuda/batch\endcsname{2.2}
\expandafter\def\csname R@xtrain/B4/pytmpar-cpu16/b100\endcsname{3.8}
\expandafter\def\csname R@xtrain/B4/pytmpar-cpu16/batch\endcsname{1.9}
\expandafter\def\csname R@xtrain/B4/tmu-cpu1/b100\endcsname{41.2}
\expandafter\def\csname R@xtrain/B4/tmu-cpu1/batch\endcsname{20.4}
\expandafter\def\csname R@xtrain/B4/tmu-cuda/b100\endcsname{24.5}
\expandafter\def\csname R@xtrain/B4/tmu-cuda/batch\endcsname{12.1}
\expandafter\def\csname R@xtrain/B5/gt-cpu16/b100\endcsname{94.1}
\expandafter\def\csname R@xtrain/B5/gt-cpu16/batch\endcsname{46.4}
\expandafter\def\csname R@xtrain/B5/pytm-cpu1/b100\endcsname{9.3}
\expandafter\def\csname R@xtrain/B5/pytm-cpu1/batch\endcsname{4.6}
\expandafter\def\csname R@xtrain/B5/pytmcuda-cuda/b100\endcsname{4.5}
\expandafter\def\csname R@xtrain/B5/pytmcuda-cuda/batch\endcsname{2.2}
\expandafter\def\csname R@xtrain/B5/pytmpar-cpu16/b100\endcsname{4.2}
\expandafter\def\csname R@xtrain/B5/pytmpar-cpu16/batch\endcsname{2.1}
\expandafter\def\csname R@xtrain/B5/tmu-cpu1/b100\endcsname{40.9}
\expandafter\def\csname R@xtrain/B5/tmu-cpu1/batch\endcsname{20.2}
\expandafter\def\csname R@xtrain/B5/tmu-cuda/b100\endcsname{26.0}
\expandafter\def\csname R@xtrain/B5/tmu-cuda/batch\endcsname{12.8}
\expandafter\def\csname R@xtrain/B7/pytm-cpu1/b100\endcsname{88.8}
\expandafter\def\csname R@xtrain/B7/pytm-cpu1/batch\endcsname{76.0}
\expandafter\def\csname R@xtrain/B7/pytmcuda-cuda/b100\endcsname{2.4}
\expandafter\def\csname R@xtrain/B7/pytmcuda-cuda/batch\endcsname{2.0}
\expandafter\def\csname R@xtrain/B7/pytmpar-cpu16/b100\endcsname{7.6}
\expandafter\def\csname R@xtrain/B7/pytmpar-cpu16/batch\endcsname{6.5}
\expandafter\def\csname R@xtrain/B7/tmu-cpu1/b100\endcsname{69.3}
\expandafter\def\csname R@xtrain/B7/tmu-cpu1/batch\endcsname{59.4}
\expandafter\def\csname R@xtrain/B7/tmu-cuda/b100\endcsname{10.3}
\expandafter\def\csname R@xtrain/B7/tmu-cuda/batch\endcsname{8.9}
\expandafter\def\csname R@xtrain/B8/pytmcuda-cuda/b100\endcsname{2.4}
\expandafter\def\csname R@xtrain/B8/pytmcuda-cuda/batch\endcsname{2.1}
\expandafter\def\csname R@xtrain/B8/tmu-cuda/b100\endcsname{12.1}
\expandafter\def\csname R@xtrain/B8/tmu-cuda/batch\endcsname{10.3}
\expandafter\def\csname R@xtrain/B9/pytmcuda-cuda/b100\endcsname{10.9}
\expandafter\def\csname R@xtrain/B9/pytmcuda-cuda/batch\endcsname{5.2}
\expandafter\def\csname R@xtrain/B9/tmu-cuda/b100\endcsname{26.2}
\expandafter\def\csname R@xtrain/B9/tmu-cuda/batch\endcsname{12.5}

\providecommand{\R}[1]{\ifcsname R@#1\endcsname\csname R@#1\endcsname\else\textcolor{red}{??#1}\fi}
\expandafter\def\csname R@bcs/conv/b1024/max\endcsname{98.96}
\expandafter\def\csname R@bcs/conv/b1024/min\endcsname{98.44}
\expandafter\def\csname R@bcs/conv/b256/max\endcsname{99.10}
\expandafter\def\csname R@bcs/conv/b256/min\endcsname{98.82}
\expandafter\def\csname R@bcs/conv/hours/b1024\endcsname{14.1}
\expandafter\def\csname R@bcs/conv/hours/b16\endcsname{1.75}
\expandafter\def\csname R@bcs/conv/lastgap\endcsname{0.19}
\expandafter\def\csname R@bcs/conv/small/max\endcsname{99.29}
\expandafter\def\csname R@bcs/conv/small/min\endcsname{99.02}
\expandafter\def\csname R@bcs/conv3/b1024/max\endcsname{97.94}
\expandafter\def\csname R@bcs/conv3/b1024/min\endcsname{96.75}
\expandafter\def\csname R@bcs/conv3/b256/max\endcsname{98.46}
\expandafter\def\csname R@bcs/conv3/b256/min\endcsname{97.65}
\expandafter\def\csname R@bcs/conv3/hours/b1024\endcsname{5.47}
\expandafter\def\csname R@bcs/conv3/hours/b32\endcsname{1.46}
\expandafter\def\csname R@bcs/conv3/lastgap\endcsname{0.30}
\expandafter\def\csname R@bcs/conv3/small/max\endcsname{99.12}
\expandafter\def\csname R@bcs/conv3/small/min\endcsname{98.83}
\expandafter\def\csname R@bcs/flat/b1024/lastmax\endcsname{79.10}
\expandafter\def\csname R@bcs/flat/b1024/lastmin\endcsname{57.31}
\expandafter\def\csname R@bcs/flat/b1024/max\endcsname{81.10}
\expandafter\def\csname R@bcs/flat/b1024/min\endcsname{70.95}
\expandafter\def\csname R@bcs/flat/hours/b1\endcsname{13.9}
\expandafter\def\csname R@bcs/flat/hours/b16\endcsname{1.35}
\expandafter\def\csname R@bcs/flat/lastgap\endcsname{1.72}
\expandafter\def\csname R@bcs/flat/small/max\endcsname{98.11}
\expandafter\def\csname R@bcs/flat/small/min\endcsname{96.02}
\expandafter\def\csname R@bcs/runs\endcsname{128}
\expandafter\def\csname R@bcs/version\endcsname{0.7.1}

\newcommand{\tabsize}{\scriptsize}

\lstdefinestyle{py}{language=Python, basicstyle=\ttfamily\scriptsize,
  keywordstyle=\color{blue!60!black}, commentstyle=\color{gray}, stringstyle=\color{teal!70!black},
  columns=fullflexible, keepspaces=true, showstringspaces=false, frame=lines,
  aboveskip=0.6em, belowskip=0.2em, xleftmargin=0.5em}

\setcopyright{none}

\begin{document}

\title{torchtsetlin: Fast, Interpretable Tsetlin Machines for PyTorch}

\author{Vajira Thambawita \orcidlink{0000-0001-6026-0929}}
\affiliation{%
  \institution{SimulaMet}
  \country{Norway}}

\author{Pål Halvorsen \orcidlink{0000-0003-2073-7029}} 
\affiliation{%
  \institution{SimulaMet}
  \country{Norway}}

\author{Ole-Christoffer Granmo \orcidlink{0000-0002-7287-030X}}
\affiliation{
  \institution{University of Agder}
  \country{Norway}}

\begin{abstract}
The Tsetlin machine learns propositional rules with teams of Tsetlin automata that are driven
by reward and penalty rather than gradients, so its models are interpretable by construction and
cheap to evaluate. Its software, however, consists of C and CUDA kernels behind
scikit-learn-style wrappers that sit outside the deep-learning ecosystem. \lib\ implements
the Tsetlin machine family as ordinary PyTorch modules: vanilla and weighted, coalesced,
regression, convolutional over sequences, images and volumes, and dense per-pixel segmentation.
The automata are a tensor buffer, so a model moves with \code{.to(device)}, is saved with
\code{state\_dict()} and learns from a mini-batch with one \code{update(x, y)} call. Its core
technique is a batched formulation of Type~I and Type~II feedback, in which automaton
transitions become event counts realised as binomial draws, with one commit per mini-batch.
On ten benchmarks taken from the existing libraries it matches their accuracy, and on all but
the two smallest it trains \R{xtrain/B7/pytmcuda-cuda/b100}--\R{xtrain/B9/pytmcuda-cuda/b100}$\times$
faster than PyTsetlinMachineCUDA and \R{xtrain/B7/tmu-cuda/b100}--\R{xtrain/B3/tmu-cuda/b100}$\times$
faster than tmu on the same GPU. The package is pure Python, MIT-licensed
and on PyPI; every number in this paper is reproduced by one script.
Code: \url{https://github.com/vlbthambawita/torchtsetlin}; package:
\url{https://pypi.org/project/torchtsetlin/}.
\end{abstract}


\maketitle

\section{Introduction}
\label{sec:intro}

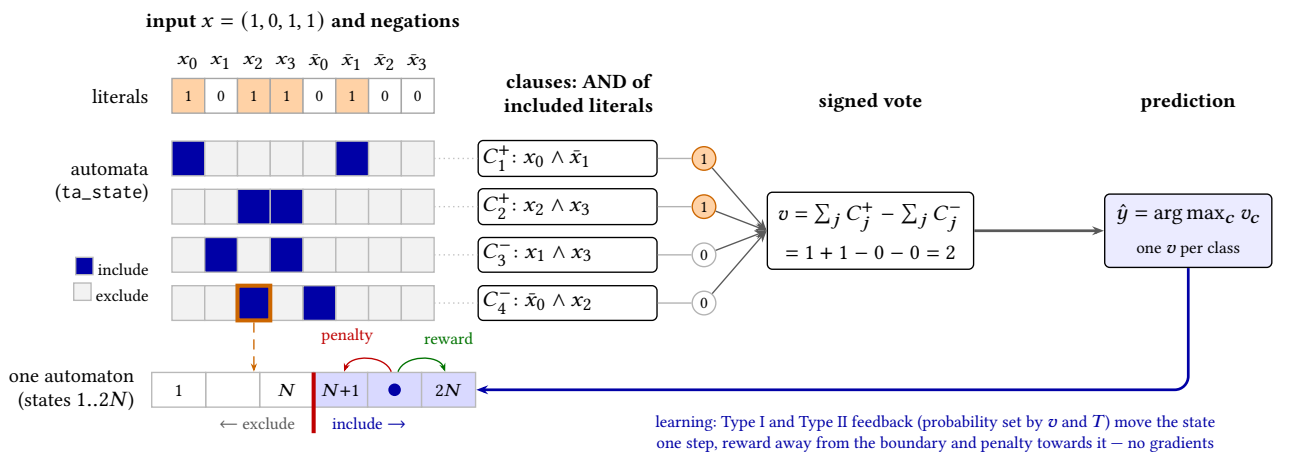
\begin{figure}[t]
\centering
\resizebox{\columnwidth}{!}{\begin{tikzpicture}[font=\scriptsize, x=1cm, y=1cm,
  cell/.style={draw=gray!55, minimum width=0.34cm, minimum height=0.36cm, inner sep=0pt, anchor=center},
  inc/.style={cell, fill=blue!65!black, text=white},
  exc/.style={cell, fill=gray!10},
  lit/.style={cell, draw=gray!70, fill=white, font=\tiny},
  lit1/.style={cell, draw=gray!70, fill=orange!35, font=\tiny},
  st/.style={cell, minimum width=0.56cm, font=\tiny},
  cbox/.style={draw, rounded corners=1.5pt, minimum height=0.36cm, inner sep=1.5pt, anchor=west, fill=white},
  bigbox/.style={draw, rounded corners=1.5pt, inner sep=3pt, anchor=west, align=center},
  arr/.style={-{Stealth[length=3.5pt]}, thick, gray!70!black},
  hdr/.style={font=\scriptsize\bfseries, anchor=south, align=center},
  side/.style={anchor=east, align=right, font=\scriptsize}]

  \node[hdr] at (1.19,2.05) {input $x=(1,0,1,1)$ and negations};
  \foreach \k/\l in {0/x_0,1/x_1,2/x_2,3/x_3,4/\bar x_0,5/\bar x_1,6/\bar x_2,7/\bar x_3}{
    \node[font=\tiny] at (\k*0.34,1.85) {$\l$};
  }
  \foreach \v [count=\k from 0] in {1,0,1,1,0,1,0,0}{
    \ifnum\v=1 \node[lit1] at (\k*0.34,1.5) {1}; \else \node[lit] at (\k*0.34,1.5) {0}; \fi
  }
  \node[side] at (-0.3,1.5) {literals};

  \foreach \row/\incl in {0/{0,5}, 1/{2,3}, 2/{1,3}, 3/{2,4}}{
    \pgfmathsetmacro{\yy}{0.85-\row*0.5}
    \foreach \k in {0,...,7}{
      \def\isin{0}
      \foreach \i in \incl {\ifnum\i=\k \xdef\isin{1}\fi}
      \ifnum\isin=1 \node[inc] at (\k*0.34,\yy) {}; \else \node[exc] at (\k*0.34,\yy) {}; \fi
    }
  }
  \node[side] at (-0.3,0.6) {automata\\(\code{ta\_state})};
  \node[side, font=\tiny] at (-0.3,-0.4)
    {\tikz\node[inc, minimum width=0.18cm, minimum height=0.18cm]{}; include\\[1pt]
     \tikz\node[exc, minimum width=0.18cm, minimum height=0.18cm]{}; exclude};

  \node[draw=orange!80!black, very thick, minimum width=0.34cm, minimum height=0.36cm, inner sep=0pt]
    (zc) at (0.68,-0.65) {};
  \foreach \i/\lbl in {0/1, 1/{}, 2/N, 3/N{+}1, 4/{}, 5/2N}{
    \pgfmathsetmacro{\xx}{-0.1+\i*0.56}
    \ifnum\i<3 \node[st] (s\i) at (\xx,-1.55) {$\lbl$};
    \else      \node[st, fill=blue!15] (s\i) at (\xx,-1.55) {$\lbl$}; \fi
  }
  \draw[orange!80!black, densely dashed, -{Stealth[length=2.5pt]}] (zc.south) -- (zc.south |- s1.north);
  \draw[red!75!black, very thick] (1.30,-1.37) -- (1.30,-2.0);
  \node[font=\tiny, text=gray!70!black, anchor=east] at (1.22,-1.9) {$\leftarrow$ exclude};
  \node[font=\tiny, text=blue!65!black, anchor=west] at (1.38,-1.9) {include $\rightarrow$};
  \fill[blue!65!black] (s4) circle (0.065);
  \draw[-{Stealth[length=2.5pt]}, green!45!black] ([xshift=1pt]s4.north) to[bend left=70] ([xshift=-1pt]s5.north);
  \node[font=\tiny, text=green!45!black, anchor=south, inner sep=1pt] at ([yshift=7pt]s5.north) {reward};
  \draw[-{Stealth[length=2.5pt]}, red!75!black] ([xshift=-1pt]s4.north) to[bend right=70] ([xshift=1pt]s3.north);
  \node[font=\tiny, text=red!75!black, anchor=south, inner sep=1pt] at ([xshift=2pt, yshift=7pt]s3.north) {penalty};
  \node[side] at (-0.45,-1.55) {one automaton\\(states $1..2N$)};

  \node[hdr] at (4.05,1.22) {clauses: AND of\\included literals};
  \foreach \row/\txt/\pol/\out in {0/{x_0 \wedge \bar x_1}/+/1, 1/{x_2 \wedge x_3}/+/1,
                                   2/{x_1 \wedge x_3}/-/0, 3/{\bar x_0 \wedge x_2}/-/0}{
    \pgfmathsetmacro{\yy}{0.85-\row*0.5}
    \node[cbox, text width=1.75cm] (c\row) at (3.0,\yy)
      {$C_{\the\numexpr\row+1\relax}^{\pol}$: $\txt$};
    \draw[gray!50, densely dotted] (2.55,\yy) -- (c\row.west);
    \ifnum\out=1
      \node[circle, draw=orange!80!black, fill=orange!35, inner sep=1.1pt, font=\tiny] (o\row) at (5.35,\yy) {1};
    \else
      \node[circle, draw=gray!60, fill=white, inner sep=1.1pt, font=\tiny] (o\row) at (5.35,\yy) {0};
    \fi
    \draw[gray!60] (c\row.east) -- (o\row);
  }

  \node[bigbox] (v) at (6.0,0.1)
    {\mbox{$v = \sum_j C^{+}_j - \sum_j C^{-}_j$}\\[3pt] $= 1 + 1 - 0 - 0 = 2$};
  \node[hdr] at (v.north |- 0,1.22) {signed vote};
  \foreach \row in {0,...,3}{ \draw[arr, thin] (o\row) -- (v.west); }
  \node[bigbox, fill=blue!8] (p) at (9.5,0.1)
    {$\hat y = \arg\max_c v_c$\\[3pt] \tiny one $v$ per class};
  \node[hdr] at (p.north |- 0,1.22) {prediction};
  \draw[arr] (v.east) -- (p.west);

  \draw[arr, blue!65!black, rounded corners=4pt] (p.south) |- (s5.east);
  \node[font=\tiny, text=blue!65!black, anchor=north, align=center] at (7.75,-1.7)
    {learning: Type~I and Type~II feedback (probability set by $v$ and $T$) move the state\\
     one step, reward away from the boundary and penalty towards it --- no gradients};
\end{tikzpicture}
}
\caption{A Tsetlin machine at a glance: each row of automata defines a clause, matching
  clauses vote with their polarity, and feedback moves each automaton's state across the
  include/exclude boundary between $N$ and $N{+}1$ (inset).}
\label{fig:tm}
\end{figure}

Multimedia systems increasingly run learned models where energy, latency and accountability
are tight: keyword spotting on always-on audio front-ends, analysis of medical images and
physiological signals, and classification of text and images at the edge. Deep networks
deliver accuracy there, but their decisions are hard to audit and their floating-point
inference is costly on small devices. The Tsetlin machine~\citep[TM;][]{granmo2018tsetlin}
offers a different trade-off. It represents each class by conjunctive clauses over Boolean
features and their negations (Figure~\ref{fig:tm}), so a trained model is a readable rule
set~\citep{rudin2019stop}, and inference reduces to bitwise AND, OR and counting, which has
led to TM chips and edge accelerators down to microwatt keyword
spotting~\citep{wheeldon2020asynchronous,maheshwari2023redress,tsetlinkws2025}. Convolutional
TMs~\citep{granmo2019convolutional} learn rules over image patches, and TMs have been applied to
medical text~\citep{berge2018medical}, ECG beats~\citep{pvc2023} and image
segmentation~\citep{ctmunet2025}.

Each (clause, literal) pair is a Tsetlin automaton~\citep{tsetlin1961,narendra1989learning}, a
counter that decides whether the literal is included, and learning is reward and penalty
instead of gradients. The family now includes weighted~\citep{phoulady2019weighted},
regression~\citep{abeyrathna2020regression} and coalesced~\citep{glimsdal2021coalesced}
machines, with regularisers such as clause-size budgets~\citep{abeyrathna2023clausesize} and
drop clause~\citep{sharma2023dropclause}.

The software has not kept pace with this reach. The reference libraries implement each variant
as a C or CUDA kernel behind a \code{fit}/\code{predict}
interface~\citep{software_pytsetlinmachine,software_pytsetlinmachinecuda,software_tmu,software_greentsetlin},
while multimedia pipelines (decoding, augmentation, feature extraction, GPU data loading and
the neural baselines a TM is compared with) are built on PyTorch~\citep{paszke2019pytorch}.
Bridging the two leaves four gaps: \emph{(i)}~the learned state sits in a private buffer that
tensor tools cannot inspect, mask or save; \emph{(ii)}~the device is fixed by the imported
package, so CPU and GPU runs follow different code paths; \emph{(iii)}~Booleanised data must
leave the framework as NumPy arrays; and \emph{(iv)}~a new feedback rule, output layer or input
structure means a new kernel. A fifth gap is evidence: each library reports results on its own
settings and hardware, and the only cross-library timing we know of covers one dataset, MNIST, and
omits both the accuracy and the hyperparameters~\citep{greentsetlin2024}.

This paper presents \lib,\footnote{Source code: \url{https://github.com/vlbthambawita/torchtsetlin}}\textsuperscript{,}\footnote{Package: \url{https://pypi.org/project/torchtsetlin/}} an open-source library that closes these gaps, and makes four
contributions:
\begin{itemize}
\item \textbf{TMs as PyTorch modules.} Vanilla, weighted, coalesced and regression machines
  share one learning loop and one tensor state, and each has convolutional variants over
  sequences (1-D), images (2-D) and volumes such as video clips or medical scans (3-D), plus
  dense per-pixel segmentation and a graph machine (Sections~\ref{sec:design}--\ref{sec:usage}).
\item \textbf{Batched feedback on the GPU.} Automaton transitions of a mini-batch become event
  counts drawn as binomials and committed once, with fused GPU kernels for the commit and for
  bit-packed convolution (Section~\ref{sec:batched}).
\item \textbf{A pre-registered, reproducible comparison} of six TM libraries on ten benchmarks
  that the existing libraries themselves publish, with no tuning and final-epoch accuracy
  (Sections~\ref{sec:protocol}--\ref{sec:results}). \lib\ matches their accuracy, and on all but
  the two smallest benchmarks trains \R{xtrain/B7/pytmcuda-cuda/b100}--\R{xtrain/B9/pytmcuda-cuda/b100}$\times$
  faster than the fastest existing GPU library on the same GPU.
\item \textbf{An open artifact}: MIT-licensed, on PyPI, documented, tested in CI, with one
  script that rebuilds every number in this paper (Section~\ref{sec:quality}).
\end{itemize}

\section{Design}
\label{sec:design}

\paragraph{State as tensors.}
A machine with $C$ clauses over $F$ features holds the automata of Figure~\ref{fig:tm} as an
integer buffer \code{ta\_state} of shape $(C, 2F)$, with memory depth $N$.
The counters are buffers, not parameters, so \code{to()}, \code{state\_dict()} and lazy shape
inference (as in \code{LazyLinear}) work unchanged. Learning is \code{update(x, y)}, the analogue
of \code{loss.backward(); optimizer.step()}, which returns the pre-update vote sums for logging;
the empty-clause convention (output~1 while learning, 0 while predicting) follows
\code{train()}/\code{eval()}.

\begin{figure*}[t]
\centering
\resizebox{\textwidth}{!}{\begin{tikzpicture}[x=1cm, y=1cm, font=\scriptsize,
  band/.style={draw=#1!35, fill=#1!5, rounded corners=3pt, line width=0.4pt},
  box/.style={draw=black!35, fill=white, rounded corners=1.5pt, line width=0.4pt, align=center,
              anchor=center, minimum height=0.72cm, minimum width=#1, text width=#1-0.16cm,
              inner sep=1pt, font=\tiny},
  pstep/.style={draw=blue!50!black, fill=blue!7, rounded corners=1pt, line width=0.4pt,
                inner xsep=1.6pt, inner ysep=1.2pt, minimum height=0.34cm, font=\tiny\ttfamily},
  arr/.style={-{Stealth[length=2.2pt,width=2.2pt]}, line width=0.4pt, black!55},
  num/.style={circle, draw=#1!60!black, fill=#1!60!black, text=white, inner sep=0pt,
              minimum size=0.38cm, font=\scriptsize\bfseries},
  lname/.style={anchor=west, font=\scriptsize\bfseries, inner sep=0pt},
  lsub/.style={anchor=west, font=\tiny\ttfamily, text=black!60, inner sep=0pt},
  op/.style={font=\normalsize, text=black!55}]
  \newcommand{\ics}[2]{{\small\color{#1!60!black}#2}}
  \newcommand{\isep}{\hspace{0.55em}}

  \foreach \y/\c/\n/\name/\sub in {0/orange/4/Workflow/{data, train, \ldots},
                                   -1.1/teal/3/Models/{models/*},
                                   -2.2/blue/2/Base/{models/base.py},
                                   -3.3/violet/1/Kernels/{functional, \_triton}}{
    \draw[band=\c] (0,\y-0.46) rectangle (15.85,\y+0.46);
    \node[num=\c] at (0.32,\y) {\n};
    \node[lname] at (0.62,\y+0.11) {\name};
    \node[lsub]  at (0.62,\y-0.17) {\sub};
  }
  \draw[band=black, fill=black!7] (0,-4.28) rectangle (15.85,-3.86);
  \node[font=\scriptsize] at (7.925,-4.07) {PyTorch: tensors and \code{nn.Module} buffers on CPU and CUDA devices};

  \node[box=2.48cm] at (3.74,0)  {\textbf{data}\\[1.5pt]\ics{orange}{\faThermometerHalf\isep\faRandom}};
  \node[box=2.48cm] at (6.42,0)  {\textbf{train}\\[1.5pt]\ics{orange}{\faSync\isep\faBell\isep\faLayerGroup}};
  \node[box=2.48cm] at (9.10,0)  {\textbf{metrics}\\[1.5pt]\ics{orange}{\faBullseye\isep\faDrawPolygon}};
  \node[box=2.48cm] at (11.78,0) {\textbf{interpret}\\[1.5pt]\ics{orange}{\faListUl\isep\faSearch}};
  \node[box=2.48cm] at (14.46,0) {\textbf{viz}\\[1.5pt]\ics{orange}{\faChartBar\isep\faMap}};

  \node[box=3.6cm] at (4.30,-1.1) {\textbf{output layer}\\[1.5pt]\ics{teal}{\faTag\isep\faIcon{share-alt}\isep\faChartLine}};
  \node[op] at (6.35,-1.1) {$\times$};
  \node[box=4.3cm] at (8.75,-1.1) {\textbf{input structure} (mixin)\\[1.5pt]\ics{teal}{\faGripLines\isep\faWaveSquare\isep\faImage\isep\faCube\isep\faBorderAll\isep\faProjectDiagram}};
  \node[op] at (11.15,-1.1) {$=$};
  \node[box=4.3cm] at (13.55,-1.1) {\textbf{13 public models}\\[1.5pt]\ics{teal}{\faMicrochip\isep\faSave\isep\faToggleOn}};

  \node[box=2.6cm] at (3.80,-2.2) {\textbf{\code{TsetlinMachineBase}}\\\code{ta\_state} (uint8) and clause weights as buffers};
  \node[box=8.0cm] (pipe) at (9.30,-2.2) {};
  \node[font=\tiny, anchor=north west, inner sep=0pt] at ([shift={(0.12,-0.06)}]pipe.north west)
    {\code{update(x, y)}: one mini-batch, one commit};
  \node[pstep, anchor=west] (s0) at ([shift={(0.12,-0.11)}]pipe.west) {\_encode};
  \foreach \t [count=\i from 1, evaluate=\i as \p using int(\i-1)] in
      {\_evaluate,\_votes,\_select\_feedback,\_feedback\_counts,\_commit}{
    \node[pstep, right=0.335cm of s\p] (s\i) {\t};
    \draw[arr] (s\p) -- (s\i);
  }
  \node[box=2.3cm] at (14.55,-2.2) {\textbf{CUDA graphs}\\update blocks captured on third use, then replayed};

  \node[box=4.27cm] at (4.635,-3.3)  {\textbf{\code{functional}}\\stateless tensor maths: literals, clause outputs, vote sums, event counts, binomial commit};
  \node[box=4.27cm] at (9.10,-3.3)   {\textbf{compiled kernels} (\code{torch.compile})\\row-sparse CPU commit with inverse-CDF binomials; per-example sequential commit};
  \node[box=4.27cm] at (13.565,-3.3) {\textbf{\code{\_triton}} (GPU)\\\code{fused\_commit}: one kernel per batch commit\\\code{conv\_eval}: OR over bit-packed patches};

  \foreach \base/\entries in {
    0/{orange/\faThermometerHalf/{thermometer encoding},
       orange/\faRandom/{hypervector encoding},
       orange/\faSync/{\code{Trainer} loop},
       orange/\faBell/{callbacks},
       orange/\faLayerGroup/{tensors or \code{DataLoader}},
       orange/\faBullseye/{accuracy, F1, calibration},
       orange/\faDrawPolygon/{IoU, Dice, boundary F1},
       orange/\faListUl/{rules with feature names},
       orange/\faSearch/{feature importance},
       orange/\faChartBar/{clause, automata, vote plots},
       orange/\faMap/{segmentation maps}},
    2/{teal/\faTag/{classification (opt.\ weights)},
       teal/\faIcon{share-alt}/{coalesced: shared clauses},
       teal/\faChartLine/{regression},
       teal/\faGripLines/{flat features},
       teal/\faWaveSquare/{1-D sequences},
       teal/\faImage/{2-D images},
       teal/\faCube/{3-D volumes},
       teal/\faBorderAll/{dense: one output per pixel},
       teal/\faProjectDiagram/{graphs with typed edges},
       teal/\faMicrochip/{\code{.to(device)}},
       teal/\faSave/{\code{state\_dict()}},
       teal/\faToggleOn/{\code{train()}\,/\,\code{eval()}}}}{
    \foreach \c/\icon/\lab [count=\k from 0] in \entries {
      \pgfmathtruncatemacro{\col}{mod(\k,6)}
      \pgfmathtruncatemacro{\row}{\base+\k/6}
      \node[font=\scriptsize, text=\c!60!black, inner sep=0pt] at (0.22+\col*2.64,-4.55-\row*0.28) {\icon};
      \node[font=\tiny, anchor=west, inner sep=0pt] at (0.45+\col*2.64,-4.55-\row*0.28) {\lab};
    }
  }
\end{tikzpicture}
}
\caption{\lib\ layer by layer. Models combine an output layer with an input structure; all of
  them share the learning loop of \code{TsetlinMachineBase}, whose update runs on stateless
  tensor maths, compiled CPU kernels or fused Triton GPU kernels, on top of PyTorch. Each layer builds only on the one below it. The legend names the icons, in the order of the boxes.}
\label{fig:layers}
\end{figure*}
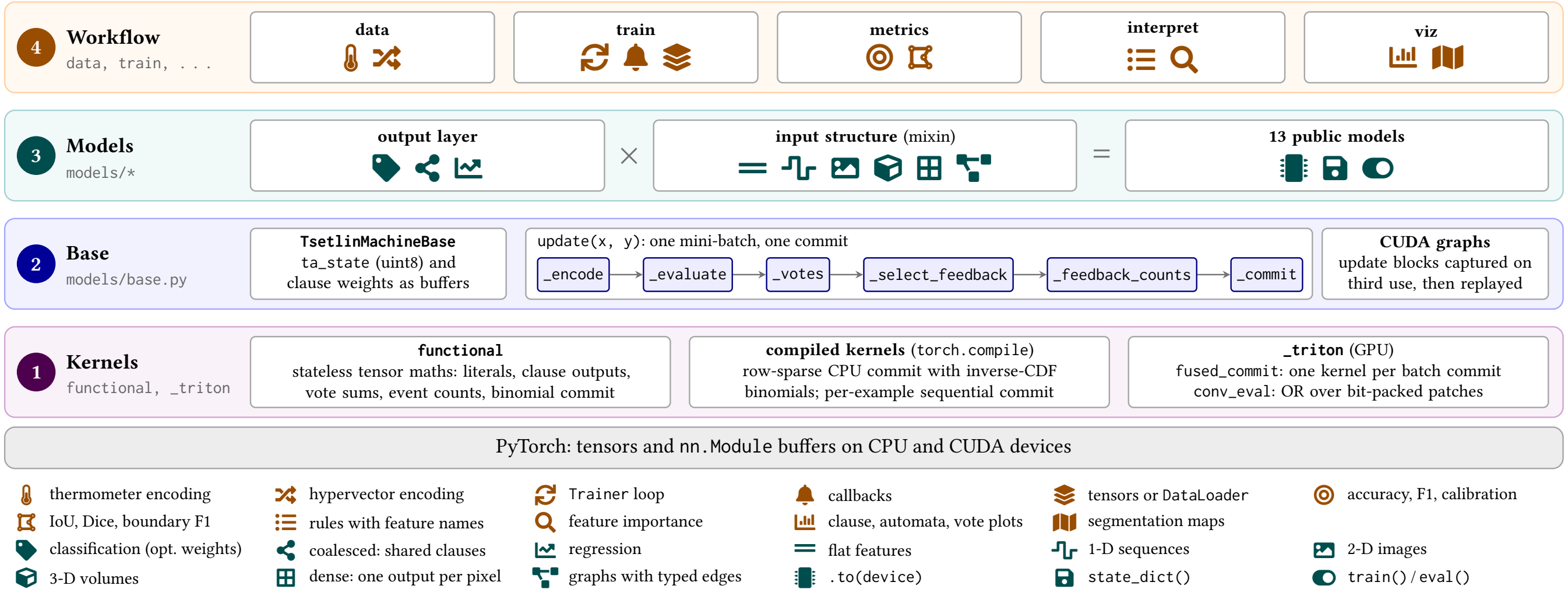

\paragraph{One learning loop, many input structures.}
The package has four layers (Figure~\ref{fig:layers}). \code{functional} holds all learning
maths as stateless tensor operations, and \code{TsetlinMachineBase} runs the generic loop. A concrete
model supplies only an output layer and a feedback policy through three hooks. A new input
structure overrides encoding, clause evaluation and feedback counting: one mixin turns every
flat model into a convolutional one, in which a clause fires if it matches any patch and
positions are thermometer-encoded~\citep{granmo2019convolutional}; a 3-D subclass changes only
the patch geometry; dense segmentation folds pixels into the batch axis so that each is a
flat example; and the graph machine~\citep{granmo2025graph} treats nodes as patches and passes
messages between clause layers along the edges.

\section{Batched Learning}
\label{sec:batched}

The classical TM learns online: an example is evaluated, its vote sums decide which clauses
receive feedback, and the automata change before the next example is
seen~\citep{granmo2018tsetlin}. On a GPU this loop leaves the device idle, because the work per
example is small and every step waits for the previous one. \lib\ learns from a mini-batch of
$B$ examples in one call, \code{update(x, y)}, in four stages that all read the automata as
they were at the start of the call; only the last stage changes them.

\paragraph{1. Evaluate.}
The batch is Booleanised into literals $L \in \{0,1\}^{B\times 2F}$, and all $C$ clauses are
evaluated at once into clause outputs $O \in \{0,1\}^{B\times C}$ and class vote sums $v$. While
learning, a clause with no included literal outputs~1. \code{update} returns these pre-update
vote sums, so training accuracy is logged without a second pass.

\paragraph{2. Select feedback.}
Each example is handled independently, as in the classical algorithm. For example $b$ with
label $y_b$, one negative class $\tilde y_b$ is drawn (uniformly, or focused by its vote sum).
Each clause of the target class is selected with probability
$(T - \mathrm{clip}(v_{b,y_b},-T,T))/2T$, each clause of the negative class with
$(T + \mathrm{clip}(v_{b,\tilde y_b},-T,T))/2T$, and the clauses of the other classes get no
feedback. Selected positive clauses of the target class and negative clauses of the negative
class receive Type~I feedback, the others Type~II. Splitting by $O$ gives
$S_{\mathrm{I}}, S_{\mathrm{II}} \in \{0,1\}^{B\times C}$, the Type~I and Type~II selections of
firing clauses, and $\bar S_{\mathrm{I}}$, the Type~I selections of non-firing clauses (Type~II
acts on firing clauses only).

\paragraph{3. Count events.}
Every elementary transition of the classical update is a Bernoulli trial on one automaton, so
the batch reduces to the number of trials per automaton, which are matrix products:
\begin{gather*}
  n^{\mathrm{true}} = S_{\mathrm{I}}^{\top} L,\quad
  n^{\mathrm{false}} = S_{\mathrm{I}}^{\top} (1-L),\\
  n^{\mathrm{II}} = S_{\mathrm{II}}^{\top} (1-L),\quad
  n^{\mathrm{Ib}} = \bar S_{\mathrm{I}}^{\top}\mathbf{1}.
\end{gather*}
Here $n^{\mathrm{true}}$ and $n^{\mathrm{false}}$ count Type~Ia events on true and false
literals of firing clauses, $n^{\mathrm{II}}$ Type~II events on false literals, and
$n^{\mathrm{Ib}}$ Type~Ib events per clause, which act on all its literals alike. A
convolutional clause fires if any patch matches; each of its events uses the literals of one
matching patch, drawn uniformly at random per example and clause, separately for Type~I and
Type~II, so $L$ holds the drawn patch for each (example, clause) pair. With weighted clauses,
the weight change of a clause is its number of Type~I minus Type~II events while firing,
$\Delta w = \mathbf{1}^{\top}(S_{\mathrm{I}} - S_{\mathrm{II}})$.

\paragraph{4. Commit.}
Trials on one automaton share one probability, so their sum is binomial, and the whole batch
is applied in one draw per automaton:
\begin{multline*}
  \Delta = \mathrm{Bin}\bigl(n^{\mathrm{true}}, p_{\mathrm{mem}}\bigr)
  - \mathrm{Bin}\bigl(n^{\mathrm{false}} + n^{\mathrm{Ib}}\mathbf{1}^{\top}, 1/s\bigr)\\
  + \min\bigl(n^{\mathrm{II}}, \max(N - \mathrm{state}, 0)\bigr),
\end{multline*}
with specificity $s$ and $p_{\mathrm{mem}} = (s-1)/s$, or $1$ with boosted true-positive
feedback. Type~II stops at the include boundary because, online, a clause stops matching an
example once it includes one of the example's false literals. States are clamped to
$[0, 2N{-}1]$ and weights to at least~1. The clause-size budget is soft, as in tmu: the Type~Ia
events of a firing clause above the budget are counted as Type~Ib, so it is forgotten instead
of reinforced. The include mask is then refreshed for the next call.

\paragraph{Memory.}
The largest intermediates have $B\times C$ entries ($B\times P\times C$ for $P$ patches). When
they exceed \code{max\_chunk\_elements}, the batch is split into chunks that add into the same
counters, so chunking bounds memory without changing the update, which is still committed once.

\paragraph{What batching changes.}
At $B=1$ the four stages are the classical update term by term. For $B>1$, every example of a
batch sees the automata and vote sums of its start. An example therefore does not see the
inclusions caused by earlier examples of the same batch, and the margin $T$, which online stops
feedback to a class once its vote sum reaches $\pm T$, reacts only once per batch, so up to $B$
pushes can land on one automaton in one commit. This is the approximation of mini-batch SGD
relative to SGD~\citep{bottou2018optimization}: small batches follow the online algorithm,
large ones overshoot. Batch size is therefore a hyperparameter, measured in
Section~\ref{sec:batching}, and \code{feedback\_mode="sequential"} keeps the exact algorithm:
it commits each example in turn inside the same call, touching only the clauses that example
selected.

\paragraph{GPU fast path.}
On CUDA, a batched update that fits in one chunk records its feedback events (the $L$ and
$S$ of stage~3) instead of counts, and one Triton kernel runs stages~3 and~4 for a tile of
(clause, literal) pairs: it computes the counts, samples the binomials, and writes the new
state and a bit-packed include mask. Convolutional
clauses are evaluated on bit-packed literals, with the OR over patches kept in registers. An
update never synchronises with the host, so its shapes are captured as CUDA graphs on their
third use and replayed; this removes the launch overhead of small models and of the sequential
mode, which runs in blocks of 64 examples. Every fused path falls back to plain PyTorch
operations, which are also the CPU path.

\section{Functionality and Usage}
\label{sec:usage}

\begin{figure*}[t]
\centering
\includegraphics[width=\textwidth]{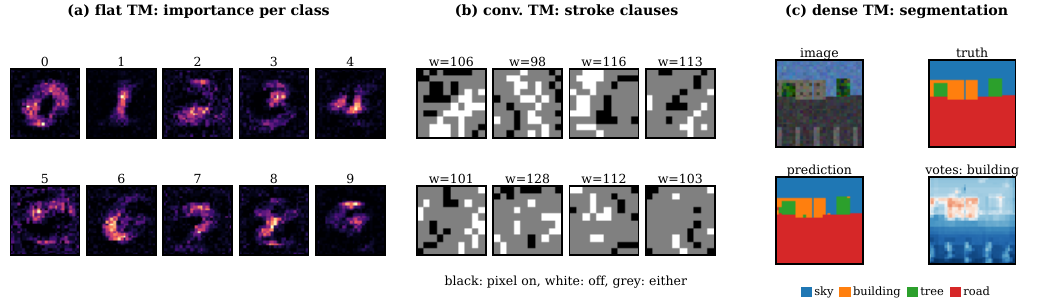}
\caption{Output of \lib's interpretation and plotting tools (\code{interpret}, \code{viz}).
  (a)~Global importance of each pixel literal for each MNIST class in a flat machine.
  (b)~Heavily weighted clauses of a convolutional MNIST machine, drawn as $10{\times}10$ patch
  templates with their weights. (c)~A dense segmentation machine on a synthetic street scene:
  input, ground truth, prediction and the vote sums for \emph{building} (red: for, blue:
  against). One seed, five to eight epochs each (\code{code/fig\_viz.py}).}
\label{fig:viz}
\end{figure*}

Listing~\ref{lst:xor} shows the complete workflow on Noisy XOR~\citep{granmo2018tsetlin}: with
40\% label noise the machine recovers and prints the generating rule. Version 0.7.3
ships \code{TsetlinMachine} (optional clause weights~\citep{phoulady2019weighted}),
\code{CoalescedTsetlinMachine} (multi-class or multi-label) and \code{RegressionTsetlinMachine},
convolutional 1-D, 2-D and 3-D variants of each, dense segmentation machines and a graph machine~\citep{granmo2025graph}. Booleanisation
encoders (thermometer, hypervector~\citep{halenka2024hypervectors} and others) are \code{nn.Module}s that run on the model's device and carry feature names
into the rules (\code{worst\_perimeter>=105.9} rather than \code{x37}). A \code{Trainer} with
callbacks accepts tensors and \code{DataLoader}s, and the interpretation tools extract rules
and global, local and per-pixel importance~\citep{blakely2020closedform}.
Figure~\ref{fig:viz} shows these tools on three small models trained with the library's own
loaders: per-class literal importance of a flat machine on MNIST, the clauses of a
convolutional machine read as image patches, and a dense machine segmenting a synthetic street
scene, with the per-pixel vote map of one class.

\begin{lstlisting}[style=py, basicstyle=\scriptsize, float=t, caption={Training, evaluating and reading a Tsetlin machine on Noisy XOR.}, label={lst:xor}]
import torchtsetlin as tt
x_train, y_train = tt.data.make_noisy_xor(5000, noise=0.4)   # Boolean features, int labels
x_test,  y_test  = tt.data.make_noisy_xor(2000, noise=0.0, seed=1)
model = tt.TsetlinMachine(n_features=12, n_classes=2, n_clauses=20, T=15, s=3.9).to("cuda")
trainer = tt.Trainer(model, batch_size=10, callbacks=[tt.EarlyStopping(patience=20)])
trainer.fit((x_train, y_train), epochs=100, val_data=(x_test, y_test))
print(trainer.evaluate((x_test, y_test)))   # {'accuracy': ...}
print(model.rules()[:2])  # ['IF x0 AND NOT x1 THEN 1', 'IF x1 AND NOT x0 THEN 1']
\end{lstlisting}

\section{Comparison with Existing Software}
\label{sec:comparison}

Table~\ref{tab:compare} compares \lib\ with the maintained TM packages.
\code{pyTsetlinMachine} \citep{software_pytsetlinmachine},
its OpenMP version \code{pyTsetlinMachineParallel} \citep{software_pytsetlinmachineparallel} and
\code{PyTsetlinMachineCUDA} \citep{software_pytsetlinmachinecuda} implement one class per
variant in C and CUDA; \code{tmu} \citep{software_tmu} unifies them over reusable clause and
weight banks with CPU and CUDA backends; \code{green-tsetlin} \citep{software_greentsetlin,greentsetlin2024}
targets efficient CPU training and deployment. Every GPU path among them is a hand-written
PyCUDA kernel. All keep the automata in a native buffer; \lib\
alone builds on a general-purpose tensor library. Below we call them pyTM, pyTM-Par, pyTM-CUDA,
tmu and GT.

\begin{table}[t]
\centering\tabsize
\setlength{\tabcolsep}{2.2pt}
\caption{Features of TM software (\cmark\ yes, \pmark\ partial, \xmark\ no). GT's 2-D
  convolution exists but fails at training through its public API. Graph machines otherwise
  ship in a separate CUDA library~\citep{granmo2025graph}.}
\label{tab:compare}
\begin{tabular}{@{}lccccc@{}}
\toprule
 & pyTM & pyTM-CUDA & tmu & GT & \lib \\
\midrule
Backend                           & C        & PyCUDA     & C\,/\,PyCUDA & C++    & PyTorch \\
CPU and GPU, one class            & \xmark   & \xmark     & \cmark     & \xmark   & \cmark \\
Conv.\ 1-D\,/\,2-D\,/\,3-D        & \xmark\cmark\xmark & \xmark\cmark\xmark & \xmark\cmark\xmark & \xmark\pmark\xmark & \cmark\cmark\cmark \\
Regression\,/\,coalesced          & \cmark\xmark & \cmark\xmark & \cmark\cmark & \xmark\cmark & \cmark\cmark \\
Dense segmentation                & \xmark   & \xmark     & \xmark     & \xmark   & \cmark \\
Graphs (message passing)          & \xmark   & \xmark     & \xmark     & \xmark   & \cmark \\
Tensor state, \code{state\_dict}  & \xmark   & \xmark     & \xmark     & \xmark   & \cmark \\
New variant without C/CUDA        & \xmark   & \xmark     & \pmark     & \xmark   & \cmark \\
\bottomrule
\end{tabular}
\end{table}

\section{Benchmark Protocol}
\label{sec:protocol}

To avoid benchmarks chosen in \lib's favour, the comparison uses only experiments that the
existing libraries already publish, in their repositories or in the papers that introduced them
(Table~\ref{tab:protocol}). Each dataset is Booleanised once and every library reads the same bits;
hyperparameters are the published ones, with no tuning, and accuracy is read at the final epoch,
never at the best one. These rules govern every comparison between libraries. Batched feedback
is a learning parameter that only \lib\ has, so Section~\ref{sec:batching} adds two studies of
the batch size in which no other library takes part; they are marked there and follow their own
protocol. Metrics are those the libraries already report: accuracy, macro F1,
precision and recall (RMSE and MAE for B10); seconds per training epoch (median of epochs
2--$E$; epoch~1 includes kernel compilation and is recorded separately); seconds to predict the
test set; peak host and GPU memory; and model size.

\begin{table*}[t]
\centering\tabsize
\setlength{\tabcolsep}{3pt}
\caption{Benchmarks, all taken from existing TM libraries or their papers. $C$ = clauses per
  class, $T$ = threshold, $s$ = specificity, w = weighted clauses. Every library implementing the
  model runs on the same Booleanised data.}
\label{tab:protocol}
\begin{tabular}{@{}l p{2.9cm} p{5.6cm} l p{5.2cm}@{}}
\toprule
ID & Dataset (task) & Booleanisation; model ($C$, $T$, $s$) & Epochs\,$\times$\,runs & Source of the experiment \\
\midrule
B1 & Noisy XOR & 12 bits, 40\% label noise; TM (10, 15, 3.9) & 200\,$\times$\,5 &
  pyTM, pyTM-Par, pyTM-CUDA demos; \citet[Tab.~15]{granmo2018tsetlin} \\
B2 & 2-D Noisy XOR & 4$\times$4 images; conv.\ TM (40, 60, 3.9), patch 2$\times$2 & 5000\,$\times$\,5 &
  pyTM, pyTM-Par demos; \citet{granmo2019convolutional} \\
B3 & Breast cancer~\citep{wolberg1995breast} & 10 bits/feature; TM (800, 40, 5.0) & 25\,$\times$\,100 splits &
  pyTM, pyTM-Par demos \\
B4 & MNIST~\citep{lecun1998mnist} & pixel $>75$; TM (2000, 50, 10) & 250\,$\times$\,5 &
  pyTM, pyTM-Par demos \\
B5 & MNIST & pixel $>75$; w-TM (2000, 5000, 10) & 100\,$\times$\,5 &
  pyTM, pyTM-Par, pyTM-CUDA demos; tmu example; \citet{phoulady2019weighted} \\
B6 & MNIST, timing & pixel $>75$; $C\in\{10,\ldots,5000\}$, $T{=}C/4$, $s{=}3$, budget 20 &
  5\,$\times$\,3 & \citet[Tab.~I]{greentsetlin2024} (cross-library); \citet{abeyrathna2021parallel} \\
B7 & MNIST & pixel $\ge 75$; conv.\ w-TM (2000, 5000, 5.0), patch 10$\times$10 & 30\,$\times$\,5 &
  pyTM, pyTM-Par, pyTM-CUDA demos; tmu example \\
B8 & Fashion-MNIST~\citep{xiao2017fashion} & adaptive Gaussian threshold; conv.\ w-TM (2000, 5000, 5.0), 10$\times$10 & 30\,$\times$\,5 &
  pyTM, pyTM-Par, pyTM-CUDA demos; \citet{granmo2019convolutional} \\
B9 & IMDb~\citep{maas2011imdb} & 5000 words, 1--2-grams, $\chi^2$ top 5000; w-TM (10000, 8000, 10) & 30\,$\times$\,5 &
  pyTM, pyTM-Par, pyTM-CUDA demos; \citet[Tab.~3]{abeyrathna2021parallel} \\
B10 & California housing~\citep{pace1997housing} (regression) & 10 bits/feature; w-RTM (1000, 5000, 2.75) & 30\,$\times$\,25 splits &
  pyTM, pyTM-Par demos; \citet[Tab.~5]{abeyrathna2021parallel} \\
\bottomrule
\end{tabular}
\end{table*}

\begin{table*}[t]
\centering\tabsize
\setlength{\tabcolsep}{4pt}
\caption{Test accuracy (\%) at the final epoch, mean $\pm$ sd over runs; B10: RMSE (lower is
  better). pyTM, pyTM-Par and GT ran once; -- not run or not supported.}
\label{tab:ml}
\begin{tabular}{@{}lrrrrrrrr@{}}
\toprule
 & \multicolumn{5}{c}{existing libraries} & \multicolumn{3}{c}{\lib\ (GPU)} \\
\cmidrule(lr){2-6}\cmidrule(l){7-9}
Bench & pyTM & Par & CUDA & tmu & GT & seq & batch & b100 \\
\midrule
B1 & 100.00 & 100.00 & 100.00{\tiny$\pm$0.00} & 100.00{\tiny$\pm$0.00} & 54.94 & 100.00{\tiny$\pm$0.00} & 100.00{\tiny$\pm$0.00} & 100.00{\tiny$\pm$0.00} \\
B2 & 98.62 & 99.97 & 98.82{\tiny$\pm$1.22} & 99.73{\tiny$\pm$0.56} & -- & 99.60{\tiny$\pm$0.46} & 99.61{\tiny$\pm$0.27} & 92.47{\tiny$\pm$3.83} \\
B3 & 99.12 & 98.25 & 97.50{\tiny$\pm$1.30} & 97.32{\tiny$\pm$1.25} & 97.37 & 97.49{\tiny$\pm$1.37} & 97.42{\tiny$\pm$1.48} & 96.86{\tiny$\pm$1.74} \\
B4 & 98.02 & 98.08 & 98.21{\tiny$\pm$0.05} & 97.97{\tiny$\pm$0.15} & 98.05 & -- & 97.90{\tiny$\pm$0.12} & 97.54{\tiny$\pm$0.13} \\
B5 & 98.12 & 98.07 & 98.11{\tiny$\pm$0.08} & 98.17{\tiny$\pm$0.12} & 10.49 & -- & 98.15{\tiny$\pm$0.08} & 98.14{\tiny$\pm$0.06} \\
B7 & 99.12 & 99.12 & 98.94{\tiny$\pm$0.04} & 99.11{\tiny$\pm$0.09} & -- & -- & 99.11{\tiny$\pm$0.05} & 99.03{\tiny$\pm$0.07} \\
B8 & -- & -- & 88.37{\tiny$\pm$2.08} & 89.97{\tiny$\pm$0.17} & -- & -- & 90.06{\tiny$\pm$0.18} & 89.70{\tiny$\pm$0.10} \\
B9 & -- & -- & 81.27{\tiny$\pm$4.45} & 88.39{\tiny$\pm$0.47} & -- & -- & 88.29{\tiny$\pm$0.93} & 87.99{\tiny$\pm$0.63} \\
B10 & -- & -- & 0.622{\tiny$\pm$0.010} & 0.614{\tiny$\pm$0.015} & -- & 0.612{\tiny$\pm$0.011} & 0.616{\tiny$\pm$0.014} & 0.621{\tiny$\pm$0.016} \\
\bottomrule
\end{tabular}

\end{table*}

\begin{table*}[t]
\centering\tabsize
\setlength{\tabcolsep}{4pt}
\caption{Training time, seconds per epoch (median of epochs 2--$E$, one exclusive run). Bold:
  the fastest GPU entry. $\times k$: threads. \lib\ on the CPU: three epochs, batch mode.}
\label{tab:train}
\begin{tabular}{@{}lrrrrrrrrrr@{}}
\toprule
 & \multicolumn{5}{c}{CPU} & \multicolumn{5}{c}{GPU (RTX 3090)} \\
\cmidrule(lr){2-6}\cmidrule(l){7-11}
 & \multicolumn{4}{c}{} & & \multicolumn{2}{c}{} & \multicolumn{3}{c}{\lib} \\
\cmidrule(l){9-11}
Bench & pyTM$\times$1 & Par$\times$16 & GT$\times$16 & tmu$\times$1 & \lib$\times$16 & CUDA & tmu & seq & batch & b100 \\
\midrule
B1 & 0.00786 & 0.00657 & 0.417 & 0.445 & -- & \textbf{0.0136} & 1.13 & 0.112 & 0.0421 & 0.0136 \\
B2 & 0.0118 & 0.00741 & -- & 0.226 & -- & \textbf{0.00884} & 0.579 & 0.102 & 0.0439 & 0.00902 \\
B3 & 0.0337 & 0.00643 & 0.0434 & 0.0766 & -- & 0.00635 & 0.125 & 0.0163 & 0.00426 & \textbf{0.00153} \\
B4 & 9.32 & 2.99 & 46.3 & 32.4 & 69.6 & 3.44 & 19.3 & -- & 1.59 & \textbf{0.788} \\
B5 & 7.44 & 3.34 & 75.5 & 32.8 & 73.0 & 3.62 & 20.8 & -- & 1.63 & \textbf{0.802} \\
B7 & 354 & 30.4 & -- & 276 & -- & 9.53 & 41.2 & -- & 4.65 & \textbf{3.99} \\
B8 & -- & -- & -- & -- & -- & 9.58 & 47.8 & -- & 4.66 & \textbf{3.97} \\
B9 & -- & -- & -- & -- & -- & 21.2 & 50.9 & -- & 4.07 & \textbf{1.94} \\
B10 & -- & -- & -- & -- & -- & 0.124 & 1.45 & 0.429 & 0.0973 & \textbf{0.0311} \\
\bottomrule
\end{tabular}

\par\medskip
\caption*{Inference time, seconds to predict the whole test set (same runs and columns).}
\begin{tabular}{@{}lrrrrrrrrrr@{}}
\toprule
 & \multicolumn{5}{c}{CPU} & \multicolumn{5}{c}{GPU (RTX 3090)} \\
\cmidrule(lr){2-6}\cmidrule(l){7-11}
 & \multicolumn{4}{c}{} & & \multicolumn{2}{c}{} & \multicolumn{3}{c}{\lib} \\
\cmidrule(l){9-11}
Bench & pyTM$\times$1 & Par$\times$16 & GT$\times$16 & tmu$\times$1 & \lib$\times$16 & CUDA & tmu & seq & batch & b100 \\
\midrule
B1 & 0.00264 & 0.00112 & 0.204 & 0.0871 & -- & \textbf{0.000570} & 0.233 & 0.000991 & 0.000752 & 0.000660 \\
B2 & 0.0211 & 0.00468 & -- & 0.205 & -- & \textbf{0.00499} & 0.488 & 0.0443 & 0.0444 & 0.0441 \\
B3 & 0.00697 & 0.00316 & 0.00554 & 0.00451 & -- & 0.000483 & 0.00666 & 0.000375 & 0.000372 & \textbf{0.000352} \\
B4 & 3.42 & 2.41 & 6.51 & 2.35 & 1.59 & 0.523 & 4.62 & -- & \textbf{0.0163} & 0.0167 \\
B5 & 3.28 & 2.45 & 4.31 & 2.40 & 1.66 & 0.550 & 4.74 & -- & \textbf{0.0164} & 0.0167 \\
B7 & 289 & 20.9 & -- & 227 & -- & 1.47 & 22.0 & -- & \textbf{0.337} & 0.340 \\
B8 & -- & -- & -- & -- & -- & 1.47 & 23.2 & -- & \textbf{0.336} & 0.337 \\
B9 & -- & -- & -- & -- & -- & 11.7 & 15.4 & -- & \textbf{0.167} & 0.168 \\
B10 & -- & -- & -- & -- & -- & 0.00220 & 0.161 & 0.000929 & 0.000746 & \textbf{0.000728} \\
\bottomrule
\end{tabular}

\end{table*}

\begin{table}[t]
\centering\tabsize
\setlength{\tabcolsep}{5pt}
\caption{B6, the protocol of \citet[Tab.~I]{greentsetlin2024}: seconds to build a model and
  train 5 epochs with a test-set prediction after each (median of 3 repeats); in brackets the
  published value (Intel i9-12900K + RTX~3090). Accuracy at $C{=}5000$, mean $\pm$ sd over the
  repeats.}
\label{tab:b6}
\begin{tabular}{@{}lrrrrrr@{}}
\toprule
Library & $C{=}10$ & $C{=}100$ & $C{=}1000$ & $C{=}2000$ & $C{=}5000$ & acc.\ (\%) \\
\midrule
pyTM CPU$\times$1 & 9.07 {\tiny(4.5)} & 20.1 {\tiny(9)} & 95.8 {\tiny(47)} & 174 {\tiny(84)} & 422 {\tiny(197)} & 96.6{\tiny$\pm$0.0} \\
pyTM-Par CPU$\times$4 & 2.25 {\tiny(1.1)} & 5.48 {\tiny(2)} & 37.1 {\tiny(17)} & 97.3 {\tiny(47)} & 262 {\tiny(149)} & 96.9{\tiny$\pm$0.2} \\
pyTM-Par CPU$\times$16 & 1.61 & 2.64 & 11.8 & 36.9 & 115 & 96.8{\tiny$\pm$0.1} \\
GT CPU$\times$1 & 1.04 {\tiny(0.3)} & 3.60 {\tiny(2)} & 35.0 {\tiny(18)} & 70.6 {\tiny(36)} & 177 {\tiny(90)} & 89.6{\tiny$\pm$0.3} \\
GT CPU$\times$4 & 12.8 {\tiny(4)} & 11.9 {\tiny(5)} & 26.0 {\tiny(11)} & 42.9 {\tiny(18)} & 97.1 {\tiny(31)} & 89.1{\tiny$\pm$0.4} \\
GT CPU$\times$16 & -- & 29.4 & 36.4 & 45.7 & 79.2 & 89.0{\tiny$\pm$0.6} \\
tmu CPU$\times$1 & 84.6 {\tiny(29.6)} & 101 {\tiny(33)} & 262 {\tiny(80)} & 432 {\tiny(128)} & 916 {\tiny(266)} & 96.3{\tiny$\pm$0.3} \\
tmu GPU & 177 {\tiny(87.8)} & 146 {\tiny(62)} & 150 {\tiny(70)} & 158 {\tiny(75)} & 186 {\tiny(95)} & 96.3{\tiny$\pm$0.2} \\
pyTM-CUDA GPU & 16.3 {\tiny(14.8)} & 27.7 {\tiny(26)} & 31.3 {\tiny(30)} & 37.7 {\tiny(36)} & 90.2 {\tiny(88)} & 90.4{\tiny$\pm$0.0} \\
\lib{} GPU [batch] & 3.61 & 3.60 & 3.63 & 3.74 & 5.39 & 96.2{\tiny$\pm$0.4} \\
\lib{} GPU [b100] & 1.93 & 1.93 & 2.02 & 2.28 & 2.99 & 96.0{\tiny$\pm$0.3} \\
\bottomrule
\end{tabular}

\end{table}

\paragraph{Libraries and hardware.}
We run pyTM \R{version/pytm} (one thread, the speed-up reference of the TM literature),
pyTM-Par \R{version/pytmpar} (16 threads), pyTM-CUDA \R{version/pytmcuda}, tmu
\R{version/tmu} (CPU and GPU), GT \R{version/gt} (16 threads) and \lib\ \R{version/tt}, each in
its own environment, on one machine: an AMD Ryzen~9 3950X (16 cores), 121\,GB RAM and an
NVIDIA RTX~3090, the GPU of GT's own comparison~\citep{greentsetlin2024}, with PyTorch~2.10,
CUDA~12.8 and Triton~3.6 for \lib. Timing runs are exclusive: one process on an otherwise idle
machine, with the load logged before and after. \lib\ runs in three modes:
\emph{seq}, the exact per-example algorithm (on the four small benchmarks only); \emph{batch},
mini-batches of 32 (16 for convolution), its documented default range; and \emph{b100}, batches
of 100, the number in pyTM-CUDA's demos. GT has no convolutional or regression model through its
public API, so it is absent from B2, B7, B8 and B10.

\paragraph{Scope.} The CPU libraries were timed on B1--B7 with one run each (seed~0); the
accuracy of the GPU libraries is the mean $\pm$ sd over five seeds (B3: 100 random splits,
B10: 25). \lib's CPU path was timed on B4 and B5 only, over three epochs (Section~\ref{sec:limits}).
The measured \lib\ is release 0.6.0 for all GPU cells except the sequential B2 cells, which were
re-run on 0.6.1 after we found that 0.6.0 predicted from a stale include cache in exactly that
configuration; 0.6.1 changes nothing else. The batch-size grid of Section~\ref{sec:batching}
(Figure~\ref{fig:batchgrid}) was measured later, on release \R{bcs/version}, whose training
code for these models differs from 0.6.1 only in how the included literals of a clause are
counted, with the same result. All
departures from the pre-registered plan are listed with the artifact
(\code{PLAN.md}, \emph{Deviations}).

\paragraph{Reproduction checks.}
Before comparing, each library ran its own published configuration once
(Table~\ref{tab:repro}). Accuracy reproduces within the pre-registered tolerance of one
percentage point wherever a published value exists. pyTM and
pyTM-CUDA train in \R{repro/R1/xtrain}$\times$ and \R{repro/R3/xtrain}$\times$ their published
time; pyTM-Par is \R{repro/R2/xtrain}$\times$ slower than its README, on a different CPU.

\begin{table}[t]
\centering\tabsize
\setlength{\tabcolsep}{3pt}
\caption{Reproduction checks: each library on its own published configuration, seed~0. In
  brackets the published value.}
\label{tab:repro}
\begin{tabular}{@{}llrr@{}}
\toprule
ID & Library and check & Acc.\ (\%) & Train s/epoch \\
\midrule
R1 & pyTM, B5 (README) & 98.16 (98.19) & 8.38 (8.8) \\
R2 & pyTM-Par$\times$10, B5 (README) & 98.16 (98.14) & 4.43 (1.7) \\
R3 & pyTM-CUDA, B8 8000 clauses (README) & 90.70 (90.94) & 42.2 (44.02) \\
R4 & tmu, B5 (example) & 98.22 & 33.1 \\
R6 & \lib\ seq, B1 \cite[Tab.~15]{granmo2018tsetlin} & 100.00 (99.30) & -- \\
\bottomrule
\end{tabular}

\end{table}

\section{Results}
\label{sec:results}

\subsection{Accuracy}
\label{sec:accuracy}

\lib\ learns what the reference libraries learn (Table~\ref{tab:ml}). In batch mode it is within
the spread of the other libraries on every benchmark but B4, where it is
\R{acc/B4/tt/batch}\% against \R{acc/B4/tmu/default}--\R{acc/B4/pytmcuda/default}\%. It reaches
\R{acc/B7/tt/batch}\% on convolutional MNIST (B7), \R{acc/B8/tt/batch}\% on Fashion-MNIST (B8,
the highest mean), \R{acc/B9/tt/batch}\% on IMDb (tmu: \R{acc/B9/tmu/default}\%), and the
sequential mode has the lowest RMSE on California housing (\R{acc/B10/tt/seq}).
Sequential and batched training agree on all four benchmarks where both run.

Three cells of the other libraries are far below their peers, and we report them as measured.
GT ends B1 at \R{acc/B1/gt/default}\% and collapses on B5 to \R{acc/B5/gt/default}\% after
reaching 97.5\% by epoch~6; its random initial automaton states and its coalesced clause pool
differ from the other libraries, and its own trainer restores the best epoch by default, which
our protocol switches off. pyTM-CUDA oscillates between epochs on IMDb after peaking near 88\%,
in every seed, so its final-epoch accuracy is \R{acc/B9/pytmcuda/default}~$\pm$~\R{sd/B9/pytmcuda/default}\%.

\subsection{Training and inference time}
\label{sec:speed}

\begin{figure}[t]
\centering
\includegraphics[width=\columnwidth]{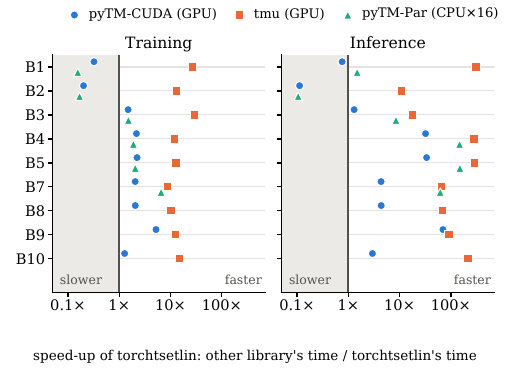}
\caption{Where \lib\ is faster: the time of each comparison library divided by that of \lib\
  (GPU, batch mode) on the same benchmark, for one training epoch and for predicting the test
  set. Right of $1\times$, \lib\ is faster. pyTM-Par is the fastest CPU library;
  B8--B10 were run on the GPU libraries only.}
\label{fig:speedup}
\end{figure}

Figure~\ref{fig:speedup} summarises where \lib\ is faster. On the GPU it trains fastest on every
benchmark with more than a few hundred automata (Table~\ref{tab:train}). Against pyTM-CUDA, b100 is
\R{xtrain/B4/pytmcuda-cuda/b100}$\times$ faster on MNIST (B4),
\R{xtrain/B7/pytmcuda-cuda/b100}$\times$ on convolutional MNIST (B7),
\R{xtrain/B9/pytmcuda-cuda/b100}$\times$ on IMDb (B9) and
\R{xtrain/B10/pytmcuda-cuda/b100}$\times$ on California housing (B10); batch mode, with
batches of 16--32, is \R{xtrain/B7/pytmcuda-cuda/batch}--\R{xtrain/B9/pytmcuda-cuda/batch}$\times$
faster on B4--B9. On B3--B10, tmu on the GPU is \R{xtrain/B7/tmu-cuda/b100}--\R{xtrain/B3/tmu-cuda/b100}$\times$
slower than b100. Against the single-threaded pyTM, the usual reference of the TM literature, a
convolutional MNIST epoch is \R{xtrain/B7/pytm-cpu1/b100}$\times$ faster (\R{train/B7/tt-cuda/b100}\,s
against \R{train/B7/pytm-cpu1/default}\,s). The two tiny models (B1, B2) are launch-bound: there b100 ties
pyTM-CUDA and the 16-thread pyTM-Par is fastest overall.

Inference shows the same pattern more strongly. \lib\ predicts the 10{,}000 MNIST test images in
\R{test/B4/tt-cuda/batch}\,s (B4), \R{xtest/B4/pytmcuda-cuda/batch}$\times$ faster than pyTM-CUDA, and
the 25{,}000 IMDb reviews \R{xtest/B9/pytmcuda-cuda/batch}$\times$ faster. Its convolutional
prediction evaluates bit-packed literals and include masks in one kernel; on the tiny B2 images,
however, pyTM-CUDA is faster (\R{test/B2/pytmcuda-cuda/default}\,s against \R{test/B2/tt-cuda/batch}\,s).

\subsection{The Green Tsetlin cross-library protocol}
\label{sec:b6}

B6 copies the only published cross-library TM timing, Table~I of \citet{greentsetlin2024}:
build a model, train five epochs and score the test set after each, for 10 to 5000 clauses.
Table~\ref{tab:b6} shows our seconds with the published ones in brackets. pyTM and
single-threaded GT take about twice their published time on our older CPU:
\R{rfive/pytm-cpu1/5000}$\times$ for pyTM and \R{rfive/gt-cpu1/5000}$\times$ for single-threaded GT at 5000 clauses, so the published
single-thread numbers reproduce up to the hardware. With four threads GT takes
\R{rfive/gt-cpu4/5000}$\times$ its published time against \R{rfive/pytmpar-cpu4/5000}$\times$ for
pyTM-Par: its thread scaling is weaker on our machine. pyTM-CUDA reproduces
closely (\R{bsix/pytmcuda-cuda/5000}\,s against 88\,s); tmu takes two to three times its
published time on both devices.
\lib\ completes the protocol at 5000 clauses in \R{bsix/tt-cuda[b100]/5000}\,s,
\R{xbsix/pytmcuda-cuda}$\times$ faster than pyTM-CUDA, \R{xbsix/tmu-cuda}$\times$ faster than
tmu on the GPU and \R{xbsix/gt-cpu16}$\times$ faster than GT on 16 threads, and its time
grows only from \R{bsix/tt-cuda[b100]/10} to \R{bsix/tt-cuda[b100]/5000}\,s across the sweep. The
accuracy column is not a like-for-like comparison: the script gives GT and the coalesced models
one pool of $C$ clauses but $C$ per class to the others, and its pyTM-CUDA model is unweighted
and has no clause budget.

\subsection{Batched versus sequential feedback}
\label{sec:batching}

Batching trades fidelity to the sequential algorithm for speed. On the four benchmarks where
both run, batch mode matches sequential mode, and b100 stays within one standard deviation
of it on B1, B3 and B10. B2 is the
exception: b100 reaches \R{acc/B2/tt/b100}\% against \R{acc/B2/tt/seq}\% for seq. A diagnostic
sweep over batch sizes (five seeds and 5000 epochs each, the first study outside the
comparison protocol) shows where fidelity breaks on this 40-clause convolutional task: batches of 16, 32 and 50 reach
\R{sweep/16}, \R{sweep/32} and \R{sweep/50}\%, the sequential result, while 100, 200 and 500
reach \R{sweep/100}, \R{sweep/200} and \R{sweep/500}\%. Batch
size is therefore a model hyperparameter in \lib, and the documentation recommends 16--64.
Sequential mode stays available and costs \R{train/B2/tt-cuda/seq}\,s per B2 epoch, against
\R{train/B2/tt-cuda/batch}\,s in batch mode.

\begin{figure*}[t]
\centering
\includegraphics[width=\textwidth]{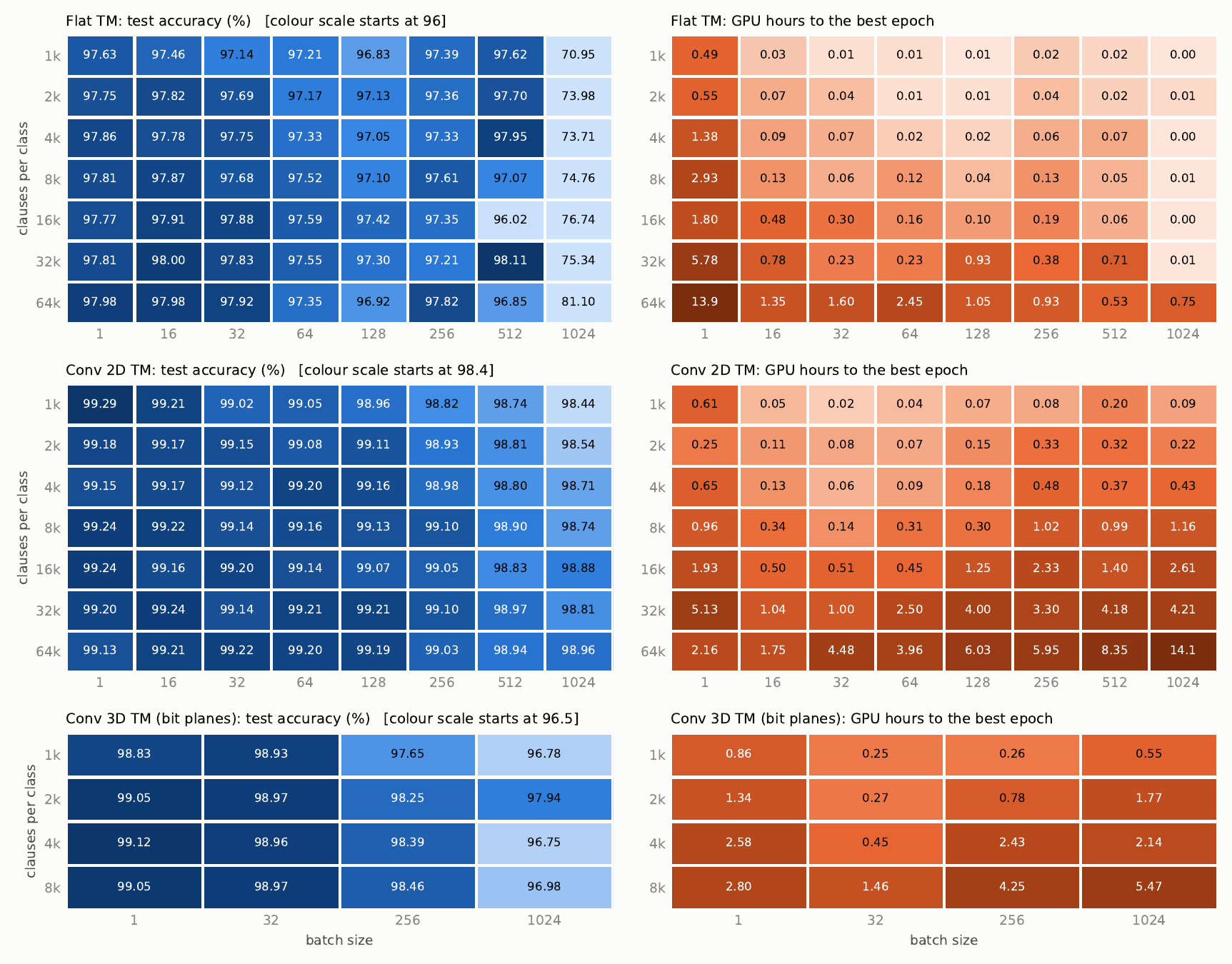}
\caption{Batch size against clauses per class on MNIST, for the flat model of B4, the
  convolutional model of B7 and a 3-D convolutional model over the eight bit planes of each
  pixel. Left: test accuracy at the best validation epoch. Right: GPU hours to reach that epoch
  (logarithmic colour scale). One seed per cell, $T$ scaled with the clause count, early
  stopping with a patience of 25 epochs; darker is higher. The accuracy colour scales are
  clipped at the stated value.}
\label{fig:batchgrid}
\end{figure*}

Figure~\ref{fig:batchgrid} repeats the question at realistic model sizes, as the second
study outside the comparison protocol of Section~\ref{sec:protocol}. It extends two of the
published benchmarks along the one axis that is new in \lib: the flat model is that of B4 and
the convolutional model that of B7, their published setting is the row with 2000 clauses,
and a batch of one updates after every example, as the reference libraries do. The 3-D
convolutional model reads the eight bit planes of each pixel. The grid covers batch sizes 1
to 1024 and 1000 to 64{,}000 clauses per class (\R{bcs/runs} runs). Its protocol differs from
the comparison in three ways: training uses 50{,}000 images and stops when the accuracy on
the remaining 10{,}000 has not improved for 25 epochs, since a fixed number of epochs would
give a large batch fewer updates; the reported value is the test accuracy at the best
validation epoch; and each cell is one seed, so differences of a few tenths of a point
between neighbouring cells are within the noise. Reading the final epoch instead, as the
comparison does, changes no convolutional cell by more than \R{bcs/conv/lastgap} points and
no 3-D cell by more than \R{bcs/conv3/lastgap}; the flat model is less stable at its last
epoch (up to \R{bcs/flat/lastgap} points lower for batches up to 512, and
\R{bcs/flat/b1024/lastmin}--\R{bcs/flat/b1024/lastmax}\% at 1024), which leaves the
conclusions below unchanged.

The convolutional model reaches \R{bcs/conv/small/min}--\R{bcs/conv/small/max}\% at every
clause count for batches up to 64, \R{bcs/conv/b256/min}--\R{bcs/conv/b256/max}\% at 256 and
\R{bcs/conv/b1024/min}--\R{bcs/conv/b1024/max}\% at 1024. The 3-D model degrades faster, from
\R{bcs/conv3/small/min}--\R{bcs/conv3/small/max}\% at batches 1 and 32 to
\R{bcs/conv3/b256/min}--\R{bcs/conv3/b256/max}\% at 256 and
\R{bcs/conv3/b1024/min}--\R{bcs/conv3/b1024/max}\% at 1024. The flat model shows no clear trend
up to a batch of 512 (\R{bcs/flat/small/min}--\R{bcs/flat/small/max}\%) and fails at 1024
(\R{bcs/flat/b1024/min}--\R{bcs/flat/b1024/max}\%).

A large batch does not pay for itself in time either, because it needs more epochs to
converge. With 64{,}000 clauses the convolutional model takes \R{bcs/conv/hours/b1024} GPU hours
to its best epoch at batch 1024, against \R{bcs/conv/hours/b16} at batch 16. At the other end,
single-example batches are the slowest way to the same accuracy: the flat model with 64{,}000
clauses needs \R{bcs/flat/hours/b1} hours at batch 1 and \R{bcs/flat/hours/b16} at batch 16.
Batches of 16 to 64 are thus both the most accurate and the cheapest setting on all three
models, which is the range the documentation recommends.

\subsection{Memory and model size}
The learned state takes the same space as in the C libraries, one byte per automaton
(\R{statemib/B4/tt-cuda/batch}\,MiB for B4, against \R{statemib/B4/pytm-cpu1/default}\,MiB for pyTM).
\lib\ needs \R{gpumib/B2/tt-cuda/b100}--\R{gpumib/B5/tt-cuda/b100}\,MiB of
GPU memory on B1--B8 and \R{gpumib/B9/tt-cuda/batch}\,MiB on IMDb, whose 20{,}000 clauses over
10{,}000 literals make the largest model here; the PyCUDA libraries hold
\R{gpumib/B4/pytmcuda-cuda/default}--\R{gpumib/B7/pytmcuda-cuda/default}\,MiB on every benchmark.
The PyTorch and CUDA runtime costs host memory: \lib\ peaks at
\R{rss/B2/tt-cuda/b100}--\R{rss/B9/tt-cuda/b100}\,GB on the GPU runs, against
\R{rss/B1/pytm-cpu1/default}--\R{rss/B9/pytmcuda-cuda/default}\,GB for the other libraries.

\subsection{Limitations}
\label{sec:limits}
\lib\ on the CPU is slow to train: \R{train/B4/tt-cpu16/batch}\,s per B4 epoch on 16 threads,
against \R{train/B4/pytm-cpu1/default}\,s for single-threaded pyTM, because dense tensor
operations on small batches cannot use the bit-level tricks of the C kernels. Its CPU
inference is the fastest of all CPU libraries (\R{test/B4/tt-cpu16/batch}\,s on B4). The
comparison uses one machine, and the CPU libraries ran once.

\section{Software Quality and Availability}
\label{sec:quality}

The suite has 257 tests, run on the CPU and on CUDA when present, and pins easily broken
behaviour such as empty-clause semantics and the absence of host synchronisation in a CUDA
update. CI tests Python 3.9, 3.11 and 3.12 on Linux; releases reach PyPI by Trusted Publishing.
The documentation (\url{https://vlbthambawita.github.io/torchtsetlin/}) has guides, examples, an API
reference, per-version benchmarks and a catalogue of 3{,}295 results from 55 TM papers.
Development is open on GitHub under the MIT licence.
\code{experiments/paper\_experiments} holds the adapters for every library, the frozen
configurations and \code{reproduce.sh}, which rebuilds every table of this paper from the JSON
records of the runs.

\begin{acks}
This work is part of the European project SEARCH, which is supported by the Innovative Health
Initiative Joint Undertaking (IHI JU) under grant agreement No.~101172997. 
\end{acks}

\bibliographystyle{ACM-Reference-Format}
\bibliography{references}


\begin{thebibliography}{33}


\ifx \showCODEN    \undefined \def \showCODEN     #1{\unskip}     \fi
\ifx \showISBNx    \undefined \def \showISBNx     #1{\unskip}     \fi
\ifx \showISBNxiii \undefined \def \showISBNxiii  #1{\unskip}     \fi
\ifx \showISSN     \undefined \def \showISSN      #1{\unskip}     \fi
\ifx \showLCCN     \undefined \def \showLCCN      #1{\unskip}     \fi
\ifx \shownote     \undefined \def \shownote      #1{#1}          \fi
\ifx \showarticletitle \undefined \def \showarticletitle #1{#1}   \fi
\ifx \showURL      \undefined \def \showURL       {\relax}        \fi
\providecommand\bibfield[2]{#2}
\providecommand\bibinfo[2]{#2}
\providecommand\natexlab[1]{#1}
\providecommand\showeprint[2][]{arXiv:#2}
\makeatletter
\@ifundefined{NAT@parse@date}{}{\let\NAT@parse@date@orig\NAT@parse@date}
\@ifundefined{NAT@parse@date}{}{\def\NAT@parse@date#1#2#3#4#5#6@@{\NAT@parse@date@orig#1#2#3#4#5#6@@\def\NAT@tempyear{0000}\def\NAT@tempexlab{{?}}\ifx\NAT@year\NAT@tempyear\ifx\NAT@exlab\NAT@tempexlab\def\NAT@date{[n.\,d.]}\else\edef\NAT@date{[n.\,d.]\NAT@exlab}\fi\fi}}
\makeatother

\bibitem[Abeyrathna et~al\mbox{.}(2023)]%
        {abeyrathna2023clausesize}
\bibfield{author}{\bibinfo{person}{K.~Darshana Abeyrathna},
  \bibinfo{person}{Ahmed A.~O. Abouzeid}, \bibinfo{person}{Bimal Bhattarai},
  \bibinfo{person}{Charul Giri}, \bibinfo{person}{Sondre Glimsdal},
  \bibinfo{person}{Ole-Christoffer Granmo}, \bibinfo{person}{Lei Jiao},
  \bibinfo{person}{Rupsa Saha}, \bibinfo{person}{Jivitesh Sharma},
  \bibinfo{person}{Svein~A. Tunheim}, {and} \bibinfo{person}{Xuan Zhang}.}
  \bibinfo{year}{2023}\natexlab{}.
\newblock \showarticletitle{Building Concise Logical Patterns by Constraining
  {Tsetlin} Machine Clause Size}. In \bibinfo{booktitle}{\emph{Proceedings of
  the Thirty-Second International Joint Conference on Artificial Intelligence
  (IJCAI)}}. \bibinfo{pages}{3395--3403}.
\newblock
\href{https://doi.org/10.24963/ijcai.2023/378}{doi:\nolinkurl{10.24963/ijcai.2023/378}}


\bibitem[Abeyrathna et~al\mbox{.}(2021)]%
        {abeyrathna2021parallel}
\bibfield{author}{\bibinfo{person}{K.~Darshana Abeyrathna},
  \bibinfo{person}{Bimal Bhattarai}, \bibinfo{person}{Morten Goodwin},
  \bibinfo{person}{Saeed~Rahimi Gorji}, \bibinfo{person}{Ole-Christoffer
  Granmo}, \bibinfo{person}{Lei Jiao}, \bibinfo{person}{Rupsa Saha}, {and}
  \bibinfo{person}{Rohan~K. Yadav}.} \bibinfo{year}{2021}\natexlab{}.
\newblock \showarticletitle{Massively Parallel and Asynchronous {Tsetlin}
  Machine Architecture Supporting Almost Constant-Time Scaling}. In
  \bibinfo{booktitle}{\emph{Proceedings of the 38th International Conference on
  Machine Learning (ICML)}} \emph{(\bibinfo{series}{Proceedings of Machine
  Learning Research}, Vol.~\bibinfo{volume}{139})}. \bibinfo{publisher}{PMLR},
  \bibinfo{pages}{10--20}.
\newblock
\urldef\tempurl%
\url{https://proceedings.mlr.press/v139/abeyrathna21a.html}
\showURL{%
\tempurl}


\bibitem[Abeyrathna et~al\mbox{.}(2020)]%
        {abeyrathna2020regression}
\bibfield{author}{\bibinfo{person}{K.~Darshana Abeyrathna},
  \bibinfo{person}{Ole-Christoffer Granmo}, \bibinfo{person}{Xuan Zhang},
  \bibinfo{person}{Lei Jiao}, {and} \bibinfo{person}{Morten Goodwin}.}
  \bibinfo{year}{2020}\natexlab{}.
\newblock \showarticletitle{The Regression {Tsetlin} Machine: A Novel Approach
  to Interpretable Nonlinear Regression}.
\newblock \bibinfo{journal}{\emph{Philosophical Transactions of the Royal
  Society A: Mathematical, Physical and Engineering Sciences}}
  \bibinfo{volume}{378}, \bibinfo{number}{2164} (\bibinfo{year}{2020}),
  \bibinfo{pages}{20190165}.
\newblock
\href{https://doi.org/10.1098/rsta.2019.0165}{doi:\nolinkurl{10.1098/rsta.2019.0165}}


\bibitem[Berge et~al\mbox{.}(2019)]%
        {berge2018medical}
\bibfield{author}{\bibinfo{person}{Geir~Thore Berge},
  \bibinfo{person}{Ole-Christoffer Granmo}, \bibinfo{person}{Tor~Oddbj{\o}rn
  Tveit}, \bibinfo{person}{Morten Goodwin}, \bibinfo{person}{Lei Jiao}, {and}
  \bibinfo{person}{Bernt~Viggo Matheussen}.} \bibinfo{year}{2019}\natexlab{}.
\newblock \showarticletitle{Using the {Tsetlin} Machine to Learn
  Human-Interpretable Rules for High-Accuracy Text Categorization With Medical
  Applications}.
\newblock \bibinfo{journal}{\emph{IEEE Access}}  \bibinfo{volume}{7}
  (\bibinfo{year}{2019}), \bibinfo{pages}{115134--115146}.
\newblock
\href{https://doi.org/10.1109/ACCESS.2019.2935416}{doi:\nolinkurl{10.1109/ACCESS.2019.2935416}}


\bibitem[Blakely and Granmo(2021)]%
        {blakely2020closedform}
\bibfield{author}{\bibinfo{person}{Christian~D. Blakely} {and}
  \bibinfo{person}{Ole-Christoffer Granmo}.} \bibinfo{year}{2021}\natexlab{}.
\newblock \showarticletitle{Closed-Form Expressions for Global and Local
  Interpretation of {Tsetlin} Machines}. In \bibinfo{booktitle}{\emph{Advances
  and Trends in Artificial Intelligence. Artificial Intelligence Practices
  (IEA/AIE 2021)}} \emph{(\bibinfo{series}{Lecture Notes in Computer Science},
  Vol.~\bibinfo{volume}{12798})}. \bibinfo{publisher}{Springer},
  \bibinfo{pages}{158--172}.
\newblock
\href{https://doi.org/10.1007/978-3-030-79457-6_14}{doi:\nolinkurl{10.1007/978-3-030-79457-6_14}}


\bibitem[Bottou et~al\mbox{.}(2018)]%
        {bottou2018optimization}
\bibfield{author}{\bibinfo{person}{L{\'e}on Bottou}, \bibinfo{person}{Frank~E.
  Curtis}, {and} \bibinfo{person}{Jorge Nocedal}.}
  \bibinfo{year}{2018}\natexlab{}.
\newblock \showarticletitle{Optimization Methods for Large-Scale Machine
  Learning}.
\newblock \bibinfo{journal}{\emph{SIAM Rev.}} \bibinfo{volume}{60},
  \bibinfo{number}{2} (\bibinfo{year}{2018}), \bibinfo{pages}{223--311}.
\newblock
\href{https://doi.org/10.1137/16M1080173}{doi:\nolinkurl{10.1137/16M1080173}}


\bibitem[Glimsdal(2025)]%
        {software_greentsetlin}
\bibfield{author}{\bibinfo{person}{Sondre Glimsdal}.}
  \bibinfo{year}{2025}\natexlab{}.
\newblock \bibinfo{title}{green-tsetlin: A Fast {Tsetlin} Machine
  Implementation, Intended for Production}.
\newblock \bibinfo{howpublished}{\url{https://github.com/ooki/green_tsetlin}}.
\newblock
\newblock
\shownote{Version 1.1.0 on PyPI (August 2025); {C++} core with {Python}
  bindings; MIT licence}.


\bibitem[Glimsdal and Granmo(2021)]%
        {glimsdal2021coalesced}
\bibfield{author}{\bibinfo{person}{Sondre Glimsdal} {and}
  \bibinfo{person}{Ole-Christoffer Granmo}.} \bibinfo{year}{2021}\natexlab{}.
\newblock \bibinfo{title}{Coalesced Multi-Output {Tsetlin} Machines with Clause
  Sharing}.
\newblock \bibinfo{howpublished}{arXiv preprint arXiv:2108.07594}.
\newblock
\showeprint[arxiv]{2108.07594}
\href{https://doi.org/10.48550/arXiv.2108.07594}{doi:\nolinkurl{10.48550/arXiv.2108.07594}}


\bibitem[Glimsdal et~al\mbox{.}(2024)]%
        {greentsetlin2024}
\bibfield{author}{\bibinfo{person}{Sondre Glimsdal}, \bibinfo{person}{Sebastian
  {\O}stby}, \bibinfo{person}{Tobias~M. Brambo}, {and}
  \bibinfo{person}{Eirik~M. Vinje}.} \bibinfo{year}{2024}\natexlab{}.
\newblock \bibinfo{title}{Green {Tsetlin}: Redefining Efficiency in {Tsetlin}
  Machine Frameworks}.
\newblock \bibinfo{howpublished}{arXiv preprint arXiv:2405.04212}.
\newblock
\showeprint[arxiv]{2405.04212}
\href{https://doi.org/10.48550/arXiv.2405.04212}{doi:\nolinkurl{10.48550/arXiv.2405.04212}}


\bibitem[Granmo(2018)]%
        {granmo2018tsetlin}
\bibfield{author}{\bibinfo{person}{Ole-Christoffer Granmo}.}
  \bibinfo{year}{2018}\natexlab{}.
\newblock \bibinfo{title}{The {Tsetlin} Machine -- A Game Theoretic Bandit
  Driven Approach to Optimal Pattern Recognition with Propositional Logic}.
\newblock \bibinfo{howpublished}{arXiv preprint arXiv:1804.01508}.
\newblock
\showeprint[arxiv]{1804.01508}
\href{https://doi.org/10.48550/arXiv.1804.01508}{doi:\nolinkurl{10.48550/arXiv.1804.01508}}


\bibitem[Granmo(2021a)]%
        {software_pytsetlinmachinecuda}
\bibfield{author}{\bibinfo{person}{Ole-Christoffer Granmo}.}
  \bibinfo{year}{2021}\natexlab{a}.
\newblock \bibinfo{title}{{PyTsetlinMachineCUDA}: Massively Parallel and
  Asynchronous Architecture for Logic-based {AI}}.
\newblock
  \bibinfo{howpublished}{\url{https://github.com/cair/PyTsetlinMachineCUDA}}.
\newblock
\newblock
\shownote{Version 0.1.9 on PyPI (June 2021); {PyCUDA} kernels; MIT licence}.


\bibitem[Granmo(2021b)]%
        {software_pytsetlinmachineparallel}
\bibfield{author}{\bibinfo{person}{Ole-Christoffer Granmo}.}
  \bibinfo{year}{2021}\natexlab{b}.
\newblock \bibinfo{title}{{pyTsetlinMachineParallel}: Multi-threaded
  Implementation of the {Tsetlin} Machine, Convolutional {Tsetlin} Machine,
  Regression {Tsetlin} Machine, and Weighted {Tsetlin} Machine}.
\newblock
  \bibinfo{howpublished}{\url{https://github.com/cair/pyTsetlinMachineParallel}}.
\newblock
\newblock
\shownote{Version 0.2.1 on PyPI (October 2021); {C} with {OpenMP}}.


\bibitem[Granmo(2024)]%
        {software_pytsetlinmachine}
\bibfield{author}{\bibinfo{person}{Ole-Christoffer Granmo}.}
  \bibinfo{year}{2024}\natexlab{}.
\newblock \bibinfo{title}{{pyTsetlinMachine}: {Tsetlin} Machine, Convolutional,
  Regression, Weighted and Embedding {Tsetlin} Machines in {C} with a {Python}
  interface}.
\newblock
  \bibinfo{howpublished}{\url{https://github.com/cair/pyTsetlinMachine}}.
\newblock
\newblock
\shownote{Version 0.6.6 on PyPI (March 2024); MIT licence}.


\bibitem[Granmo et~al\mbox{.}(2025)]%
        {granmo2025graph}
\bibfield{author}{\bibinfo{person}{Ole-Christoffer Granmo},
  \bibinfo{person}{Youmna Abdelwahab}, \bibinfo{person}{Per-Arne Andersen},
  \bibinfo{person}{Karl Audun~K. Borgersen}, \bibinfo{person}{Paul F.~A.
  Clarke}, \bibinfo{person}{Kunal Dumbre}, \bibinfo{person}{Ylva
  Gr{\o}nnings{\ae}ter}, \bibinfo{person}{Vojtech Halenka},
  \bibinfo{person}{Runar Helin}, \bibinfo{person}{Lei Jiao},
  \bibinfo{person}{Ahmed Khalid}, \bibinfo{person}{Rebekka Omslandseter},
  \bibinfo{person}{Rupsa Saha}, \bibinfo{person}{Mayur Shende}, {and}
  \bibinfo{person}{Xuan Zhang}.} \bibinfo{year}{2025}\natexlab{}.
\newblock \bibinfo{title}{The {Tsetlin} Machine Goes Deep: Logical Learning and
  Reasoning With Graphs}.
\newblock \bibinfo{howpublished}{arXiv preprint arXiv:2507.14874}.
\newblock
\showeprint[arxiv]{2507.14874}
\href{https://doi.org/10.48550/arXiv.2507.14874}{doi:\nolinkurl{10.48550/arXiv.2507.14874}}


\bibitem[Granmo and Andersen(2024)]%
        {software_tmu}
\bibfield{author}{\bibinfo{person}{Ole-Christoffer Granmo} {and}
  \bibinfo{person}{Per-Arne Andersen}.} \bibinfo{year}{2024}\natexlab{}.
\newblock \bibinfo{title}{{TMU}: {Tsetlin} Machine Unified}.
\newblock \bibinfo{howpublished}{\url{https://github.com/cair/tmu}}.
\newblock
\newblock
\shownote{Version 0.8.3 on PyPI (March 2024); MIT licence}.


\bibitem[Granmo et~al\mbox{.}(2019)]%
        {granmo2019convolutional}
\bibfield{author}{\bibinfo{person}{Ole-Christoffer Granmo},
  \bibinfo{person}{Sondre Glimsdal}, \bibinfo{person}{Lei Jiao},
  \bibinfo{person}{Morten Goodwin}, \bibinfo{person}{Christian~W. Omlin}, {and}
  \bibinfo{person}{Geir~Thore Berge}.} \bibinfo{year}{2019}\natexlab{}.
\newblock \bibinfo{title}{The Convolutional {Tsetlin} Machine}.
\newblock \bibinfo{howpublished}{arXiv preprint arXiv:1905.09688}.
\newblock
\showeprint[arxiv]{1905.09688}
\href{https://doi.org/10.48550/arXiv.1905.09688}{doi:\nolinkurl{10.48550/arXiv.1905.09688}}


\bibitem[Halenka et~al\mbox{.}(2024)]%
        {halenka2024hypervectors}
\bibfield{author}{\bibinfo{person}{Vojtech Halenka}, \bibinfo{person}{Ahmed~K.
  Kadhim}, \bibinfo{person}{Paul F.~A. Clarke}, \bibinfo{person}{Bimal
  Bhattarai}, \bibinfo{person}{Rupsa Saha}, \bibinfo{person}{Ole-Christoffer
  Granmo}, \bibinfo{person}{Lei Jiao}, {and} \bibinfo{person}{Per-Arne
  Andersen}.} \bibinfo{year}{2024}\natexlab{}.
\newblock \showarticletitle{Exploring Effects of Hyperdimensional Vectors for
  {Tsetlin} Machines}. In \bibinfo{booktitle}{\emph{2024 International
  Symposium on the {Tsetlin} Machine (ISTM)}}. \bibinfo{publisher}{IEEE},
  \bibinfo{pages}{1--8}.
\newblock
\href{https://doi.org/10.1109/ISTM62799.2024.10931546}{doi:\nolinkurl{10.1109/ISTM62799.2024.10931546}}


\bibitem[LeCun et~al\mbox{.}(1998)]%
        {lecun1998mnist}
\bibfield{author}{\bibinfo{person}{Yann LeCun}, \bibinfo{person}{L{\'e}on
  Bottou}, \bibinfo{person}{Yoshua Bengio}, {and} \bibinfo{person}{Patrick
  Haffner}.} \bibinfo{year}{1998}\natexlab{}.
\newblock \showarticletitle{Gradient-Based Learning Applied to Document
  Recognition}.
\newblock \bibinfo{journal}{\emph{Proc. IEEE}} \bibinfo{volume}{86},
  \bibinfo{number}{11} (\bibinfo{year}{1998}), \bibinfo{pages}{2278--2324}.
\newblock
\href{https://doi.org/10.1109/5.726791}{doi:\nolinkurl{10.1109/5.726791}}


\bibitem[Lin et~al\mbox{.}(2025)]%
        {tsetlinkws2025}
\bibfield{author}{\bibinfo{person}{Baizhou Lin}, \bibinfo{person}{Yuetong
  Fang}, \bibinfo{person}{Renjing Xu}, \bibinfo{person}{Rishad Shafik}, {and}
  \bibinfo{person}{Jagmohan Chauhan}.} \bibinfo{year}{2025}\natexlab{}.
\newblock \bibinfo{title}{{TsetlinKWS}: A 65nm 16.58{$\mu$W}, 0.63mm$^2$
  State-Driven Convolutional {Tsetlin} Machine-Based Accelerator for Keyword
  Spotting}.
\newblock \bibinfo{howpublished}{arXiv preprint arXiv:2510.24282}.
\newblock
\showeprint[arxiv]{2510.24282}
\href{https://doi.org/10.48550/arXiv.2510.24282}{doi:\nolinkurl{10.48550/arXiv.2510.24282}}


\bibitem[Liu and Abeyrathna(2025)]%
        {ctmunet2025}
\bibfield{author}{\bibinfo{person}{Yi Liu} {and} \bibinfo{person}{K.~Darshana
  Abeyrathna}.} \bibinfo{year}{2025}\natexlab{}.
\newblock \showarticletitle{{CTM-UNet}: Integrating Convolutional {Tsetlin}
  Machines with {U-Net} Architecture for Semantic Segmentation}. In
  \bibinfo{booktitle}{\emph{2025 International Symposium on the {Tsetlin}
  Machine (ISTM)}}. \bibinfo{publisher}{IEEE}, \bibinfo{pages}{30--36}.
\newblock
\href{https://doi.org/10.1109/ISTM67926.2025.00013}{doi:\nolinkurl{10.1109/ISTM67926.2025.00013}}


\bibitem[Maas et~al\mbox{.}(2011)]%
        {maas2011imdb}
\bibfield{author}{\bibinfo{person}{Andrew~L. Maas}, \bibinfo{person}{Raymond~E.
  Daly}, \bibinfo{person}{Peter~T. Pham}, \bibinfo{person}{Dan Huang},
  \bibinfo{person}{Andrew~Y. Ng}, {and} \bibinfo{person}{Christopher Potts}.}
  \bibinfo{year}{2011}\natexlab{}.
\newblock \showarticletitle{Learning Word Vectors for Sentiment Analysis}. In
  \bibinfo{booktitle}{\emph{Proceedings of the 49th Annual Meeting of the
  Association for Computational Linguistics: Human Language Technologies}}.
  \bibinfo{pages}{142--150}.
\newblock


\bibitem[Maheshwari et~al\mbox{.}(2023)]%
        {maheshwari2023redress}
\bibfield{author}{\bibinfo{person}{Sidharth Maheshwari},
  \bibinfo{person}{Tousif Rahman}, \bibinfo{person}{Rishad Shafik},
  \bibinfo{person}{Alex Yakovlev}, \bibinfo{person}{Ashur Rafiev},
  \bibinfo{person}{Lei Jiao}, {and} \bibinfo{person}{Ole-Christoffer Granmo}.}
  \bibinfo{year}{2023}\natexlab{}.
\newblock \showarticletitle{{REDRESS}: Generating Compressed Models for Edge
  Inference Using {Tsetlin} Machines}.
\newblock \bibinfo{journal}{\emph{IEEE Transactions on Pattern Analysis and
  Machine Intelligence}} \bibinfo{volume}{45}, \bibinfo{number}{9}
  (\bibinfo{year}{2023}), \bibinfo{pages}{11152--11168}.
\newblock
\href{https://doi.org/10.1109/TPAMI.2023.3268415}{doi:\nolinkurl{10.1109/TPAMI.2023.3268415}}


\bibitem[Narendra and Thathachar(1989)]%
        {narendra1989learning}
\bibfield{author}{\bibinfo{person}{Kumpati~S. Narendra} {and}
  \bibinfo{person}{Mandayam A.~L. Thathachar}.}
  \bibinfo{year}{1989}\natexlab{}.
\newblock \bibinfo{booktitle}{\emph{Learning Automata: An Introduction}}.
\newblock \bibinfo{publisher}{Prentice Hall}, \bibinfo{address}{Englewood
  Cliffs, NJ}.
\newblock
\showISBNx{0-13-485558-2}


\bibitem[Pace and Barry(1997)]%
        {pace1997housing}
\bibfield{author}{\bibinfo{person}{R.~Kelley Pace} {and}
  \bibinfo{person}{Ronald Barry}.} \bibinfo{year}{1997}\natexlab{}.
\newblock \showarticletitle{Sparse Spatial Autoregressions}.
\newblock \bibinfo{journal}{\emph{Statistics \& Probability Letters}}
  \bibinfo{volume}{33}, \bibinfo{number}{3} (\bibinfo{year}{1997}),
  \bibinfo{pages}{291--297}.
\newblock
\href{https://doi.org/10.1016/S0167-7152(96)00140-X}{doi:\nolinkurl{10.1016/S0167-7152(96)00140-X}}


\bibitem[Paszke et~al\mbox{.}(2019)]%
        {paszke2019pytorch}
\bibfield{author}{\bibinfo{person}{Adam Paszke}, \bibinfo{person}{Sam Gross},
  \bibinfo{person}{Francisco Massa}, \bibinfo{person}{Adam Lerer},
  \bibinfo{person}{James Bradbury}, \bibinfo{person}{Gregory Chanan},
  \bibinfo{person}{Trevor Killeen}, \bibinfo{person}{Zeming Lin},
  \bibinfo{person}{Natalia Gimelshein}, \bibinfo{person}{Luca Antiga},
  \bibinfo{person}{Alban Desmaison}, \bibinfo{person}{Andreas K{\"o}pf},
  \bibinfo{person}{Edward Yang}, \bibinfo{person}{Zachary DeVito},
  \bibinfo{person}{Martin Raison}, \bibinfo{person}{Alykhan Tejani},
  \bibinfo{person}{Sasank Chilamkurthy}, \bibinfo{person}{Benoit Steiner},
  \bibinfo{person}{Lu Fang}, \bibinfo{person}{Junjie Bai}, {and}
  \bibinfo{person}{Soumith Chintala}.} \bibinfo{year}{2019}\natexlab{}.
\newblock \showarticletitle{{PyTorch}: An Imperative Style, High-Performance
  Deep Learning Library}. In \bibinfo{booktitle}{\emph{Advances in Neural
  Information Processing Systems 32 (NeurIPS 2019)}}.
  \bibinfo{pages}{8024--8035}.
\newblock


\bibitem[Phoulady et~al\mbox{.}(2019)]%
        {phoulady2019weighted}
\bibfield{author}{\bibinfo{person}{Adrian Phoulady},
  \bibinfo{person}{Ole-Christoffer Granmo}, \bibinfo{person}{Saeed~Rahimi
  Gorji}, {and} \bibinfo{person}{Hady~Ahmady Phoulady}.}
  \bibinfo{year}{2019}\natexlab{}.
\newblock \bibinfo{title}{The Weighted {Tsetlin} Machine: Compressed
  Representations with Weighted Clauses}.
\newblock \bibinfo{howpublished}{arXiv preprint arXiv:1911.12607}.
\newblock
\showeprint[arxiv]{1911.12607}
\href{https://doi.org/10.48550/arXiv.1911.12607}{doi:\nolinkurl{10.48550/arXiv.1911.12607}}
\newblock
\shownote{Presented at the Ninth International Workshop on Statistical
  Relational AI (StarAI 2020)}.


\bibitem[Rudin(2019)]%
        {rudin2019stop}
\bibfield{author}{\bibinfo{person}{Cynthia Rudin}.}
  \bibinfo{year}{2019}\natexlab{}.
\newblock \showarticletitle{Stop Explaining Black Box Machine Learning Models
  for High Stakes Decisions and Use Interpretable Models Instead}.
\newblock \bibinfo{journal}{\emph{Nature Machine Intelligence}}
  \bibinfo{volume}{1}, \bibinfo{number}{5} (\bibinfo{year}{2019}),
  \bibinfo{pages}{206--215}.
\newblock
\href{https://doi.org/10.1038/s42256-019-0048-x}{doi:\nolinkurl{10.1038/s42256-019-0048-x}}


\bibitem[Sharma et~al\mbox{.}(2023)]%
        {sharma2023dropclause}
\bibfield{author}{\bibinfo{person}{Jivitesh Sharma}, \bibinfo{person}{Rohan
  Yadav}, \bibinfo{person}{Ole-Christoffer Granmo}, {and} \bibinfo{person}{Lei
  Jiao}.} \bibinfo{year}{2023}\natexlab{}.
\newblock \showarticletitle{Drop Clause: Enhancing Performance, Robustness and
  Pattern Recognition Capabilities of the {Tsetlin} Machine}. In
  \bibinfo{booktitle}{\emph{Proceedings of the AAAI Conference on Artificial
  Intelligence}}, Vol.~\bibinfo{volume}{37}. \bibinfo{pages}{13547--13555}.
\newblock
\href{https://doi.org/10.1609/aaai.v37i11.26588}{doi:\nolinkurl{10.1609/aaai.v37i11.26588}}


\bibitem[Tsetlin(1961)]%
        {tsetlin1961}
\bibfield{author}{\bibinfo{person}{Michael~L. Tsetlin}.}
  \bibinfo{year}{1961}\natexlab{}.
\newblock \showarticletitle{On Behaviour of Finite Automata in Random Medium}.
\newblock \bibinfo{journal}{\emph{Avtomatika i Telemekhanika}}
  \bibinfo{volume}{22}, \bibinfo{number}{10} (\bibinfo{year}{1961}),
  \bibinfo{pages}{1345--1354}.
\newblock
\newblock
\shownote{In Russian. English translation in \emph{Automation and Remote
  Control}, 22:1210--1219, 1961}.


\bibitem[Wheeldon et~al\mbox{.}(2021)]%
        {wheeldon2020asynchronous}
\bibfield{author}{\bibinfo{person}{Adrian Wheeldon}, \bibinfo{person}{Alex
  Yakovlev}, \bibinfo{person}{Rishad Shafik}, {and} \bibinfo{person}{Jordan
  Morris}.} \bibinfo{year}{2021}\natexlab{}.
\newblock \showarticletitle{Low-Latency Asynchronous Logic Design for Inference
  at the Edge}. In \bibinfo{booktitle}{\emph{2021 Design, Automation \& Test in
  Europe Conference \& Exhibition (DATE)}}. \bibinfo{publisher}{IEEE},
  \bibinfo{pages}{370--373}.
\newblock
\href{https://doi.org/10.23919/DATE51398.2021.9474126}{doi:\nolinkurl{10.23919/DATE51398.2021.9474126}}
\newblock
\shownote{arXiv:2012.03402 (2020)}.


\bibitem[Wolberg et~al\mbox{.}(1995)]%
        {wolberg1995breast}
\bibfield{author}{\bibinfo{person}{William Wolberg}, \bibinfo{person}{Olvi
  Mangasarian}, \bibinfo{person}{Nick Street}, {and} \bibinfo{person}{W.
  Street}.} \bibinfo{year}{1995}\natexlab{}.
\newblock \bibinfo{title}{Breast Cancer {Wisconsin} ({Diagnostic})}.
\newblock \bibinfo{howpublished}{UCI Machine Learning Repository}.
\newblock
\href{https://doi.org/10.24432/C5DW2B}{doi:\nolinkurl{10.24432/C5DW2B}}


\bibitem[Xiao et~al\mbox{.}(2017)]%
        {xiao2017fashion}
\bibfield{author}{\bibinfo{person}{Han Xiao}, \bibinfo{person}{Kashif Rasul},
  {and} \bibinfo{person}{Roland Vollgraf}.} \bibinfo{year}{2017}\natexlab{}.
\newblock \bibinfo{title}{{Fashion-MNIST}: A Novel Image Dataset for
  Benchmarking Machine Learning Algorithms}.
\newblock \bibinfo{howpublished}{arXiv preprint arXiv:1708.07747}.
\newblock
\showeprint[arxiv]{1708.07747}
\href{https://doi.org/10.48550/arXiv.1708.07747}{doi:\nolinkurl{10.48550/arXiv.1708.07747}}


\bibitem[Zhang et~al\mbox{.}(2023)]%
        {pvc2023}
\bibfield{author}{\bibinfo{person}{Jinbao Zhang}, \bibinfo{person}{Xuan Zhang},
  \bibinfo{person}{Lei Jiao}, \bibinfo{person}{Ole-Christoffer Granmo},
  \bibinfo{person}{Yongjun Qian}, {and} \bibinfo{person}{Fan Pan}.}
  \bibinfo{year}{2023}\natexlab{}.
\newblock \bibinfo{title}{Interpretable {Tsetlin} Machine-based Premature
  Ventricular Contraction Identification}.
\newblock \bibinfo{howpublished}{arXiv preprint arXiv:2301.10181}.
\newblock
\showeprint[arxiv]{2301.10181}
\href{https://doi.org/10.48550/arXiv.2301.10181}{doi:\nolinkurl{10.48550/arXiv.2301.10181}}


\end{thebibliography}

\end{document}